\documentclass[AMA,Times1COL]{WileyNJDv5}

\articletype{Research Article}%

\received{00}
\revised{00}
\accepted{00}
\journal{International Journal for Numerical Methods in Engineering}
\volume{00}
\copyyear{2025}
\startpage{1}

\usepackage{algorithm}
\usepackage{graphicx,url}
\usepackage{caption}
\usepackage{subfig}
\usepackage{tabularx}
\usepackage{amsmath,amssymb,amsopn}
\usepackage{color}
\usepackage{calc}
\usepackage{longtable}
\usepackage{multirow}
\usepackage{mathrsfs}
\usepackage{array}
\usepackage{hyperref}
\usepackage{cleveref}
\usepackage{booktabs}
\usepackage[outline]{contour}

\def\fv{\vec{f}}

\def\rv{\vec{r}}

\def\uv{\vec{u}}

\def\xv{\vec{x}}
\def\yv{\vec{y}}
\def\zv{\vec{z}}
\def\thetav{\vec{\theta}}

\def\etav{\vec{\eta}}

\def\rm{\mat{r}}

\def\Am{\mat{A}}

\def\Id{\mat{I}}

\def\Km{\mat{K}}

\def\Qm{\mat{Q}}
\def\Pm{\mat{P}}
\def\Rm{\mat{R}}

\def\R{\mathbb{R}}			
\def\E {\mathbb{E}}			

\def\D{\pazocal{D}}			
\def\H{\pazocal{H}}			
\def\D{\pazocal{D}} 			

\def\A{\pazocal{A}}
\def\B{\pazocal{B}}
\def\Y{\pazocal{Y}}

\def\G{\pazocal{G}}

\def\T{\pazocal{T}}
\def\M{\pazocal{M}}
\def\X{\pazocal{X}}
\def\Y{\pazocal{Y}}
\def\G{\pazocal{G}}
\def\Er{\pazocal{E}}

\DeclareMathAlphabet{\pazocal}{OMS}{zplm}{m}{n}

\renewcommand{\vec}[1]{\boldsymbol{#1}}
\newcommand{\mat}[1]{\boldsymbol{\mathrm{#1}}}

\DeclareMathOperator*{\argmin}{arg\,min}
\def\argdot{{\hspace{0.18em}\cdot\hspace{0.18em}}}

\begin{document}

\title{Hybrid Iterative Solvers with Geometry-Aware Neural Preconditioners for Parametric PDEs}

\author[1]{Youngkyu Lee}

\author[2]{Francesc Levrero Florencio}

\author[3]{Jay Pathak}

\author[1]{George Em Karniadakis}

\authormark{Lee \textsc{et al.}}
\titlemark{Hybrid Iterative Solvers with Geometry-Aware Neural Preconditioners for Parametric PDEs}

\address[1]{
\orgdiv{Division of Applied Mathematics}, \orgname{Brown University}, \orgaddress{\state{RI}, \country{USA}}
}

\address[2]{
\orgname{Ansys UK Ltd}, 
\orgaddress{\country{UK}}
}

\address[3]{
\orgname{Ansys Inc}, 
\orgaddress{\country{USA}}
}

\corres{George Em Karniadakis, \email{george\_karniadakis@brown.edu}}

\abstract[Abstract]{The convergence behavior of classical iterative solvers for parametric partial differential equations~(PDEs) is often highly sensitive to the domain and specific discretization of PDEs. 
Previously, we introduced {\em hybrid\/} solvers by combining the classical solvers with neural operators for a specific geometry~\cite{zhang2024blending}, but they tend to under-perform in geometries not encountered during training.
To address this challenge, we introduce Geo-DeepONet, a geometry-aware deep operator network that incorporates domain information extracted from finite element discretizations.
Geo-DeepONet enables accurate operator learning across arbitrary unstructured meshes without requiring retraining.
Building on this, we develop a class of geometry-aware hybrid preconditioned iterative solvers by coupling Geo-DeepONet with traditional methods such as relaxation schemes and Krylov subspace algorithms.
Through numerical experiments on parametric PDEs posed over diverse unstructured domains, we demonstrate the enhanced robustness and efficiency of the proposed hybrid solvers for multiple real-world applications.}


\maketitle

\section{Introduction}
The numerical solutions of parametric elliptic partial differential equations~(PDEs) play an important role in many real-world scientific applications.
To obtain a high-fidelity numerical solution, which is often found with the finite element method~(FEM), it is necessary to solve a large system of linear equations~\cite{ciarlet2002finite}.
Typically, iterative solvers are used to find solutions, but their convergence is highly related to the condition number of the matrix of the linear system~\cite{saad2003iterative}.
In particular, the condition number of the system is highly dependent on the discretization of the domain.
When the domain has complex geometrical features, a fine discretization is inevitable, which causes the large condition number~\cite{brenner2008mathematical}.

To improve the convergence of iterative solvers, various approaches have been proposed by introducing a suitable preconditioner~\cite{briggs2000multigrid,meijerink1977iterative,toselli2006domain,young1954iterative}.
For example, when the geometric structure of the problem is known, domain decomposition methods~\cite{toselli2006domain} or multigrid methods~\cite{briggs2000multigrid} are commonly used to guarantee algorithmic scalability.
In contrast, preconditioners like incomplete LU decomposition~(ILU)~\cite{meijerink1977iterative} or successive over-relaxation~(SOR)~\cite{young1954iterative} are employed when geometric information is unavailable.

On the other hand, machine learning~(ML) approaches have recently been attracting attention for solving PDEs.
One of the most popular approaches are the physics-informed neural networks~(PINNs)~\cite{RAISSI2019686}, which are trained by minimizing the mean square error of the residual of the PDEs, boundary/initial conditions, and observed data.
PINN have become an alternative to classical solvers, leading to numerous applications and variations~\cite{cai2021physics2,cai2021physics,lee2026two,mao2020physics,pang2019fpinns}.
Another approach is operator learning~\cite{jin2022mionet,li2021fourier,lu2021learning}, also known as neural operators~(NOs), which makes the neural networks learn the mapping between parametric functions and their corresponding solutions.
The main advantage of NOs is their reusability, i.e., the network can predict the solutions in real time without extra training.

Recently, hybrid preconditioning strategies that utilize the pre-trained NO as the preconditioner of iterative solvers have been developed~\cite{cui2022fourier,cui2025neural,kahana2023geometry,kopanivcakova2024deeponet,kopanivcakova2025leveraging,lee2024automatic,lee2025automatic,lee2025fast,zhang2024blending}.
Classical iterative solvers are known to effectively reduce high-frequency modes but often struggle with low-frequency modes~\cite{saad2003iterative}.
In contrast, neural network-based solvers, such as NOs, show the opposite behavior caused by the phenomenon known as the spectral bias of neural networks~\cite{rahaman2019spectral}.
The main idea behind hybrid preconditioning is that we can achieve the best of both worlds by leveraging the strengths of both approaches.
The hybrid preconditioner significantly improves the convergence of iterative solvers compared to standard preconditioners in structured domains~\cite{cui2022fourier,cui2025neural,kopanivcakova2024deeponet,lee2025fast,zhang2024blending}.
However, the hybrid preconditioning strategies proposed in~\cite{kopanivcakova2024deeponet,kopanivcakova2025leveraging,lee2025fast,zhang2024blending} require that the NO is trained on the same domain.
While these strategies enable fast iterative solvers if the NO is trained on a specific domain, they limit the flexibility of the solver to new domains.
To address this challenge, a transfer learning technique was proposed in~\cite{kahana2023geometry}, but it only partially overcame the issue.

\begin{figure}
	\centering
	\includegraphics[width=\linewidth]{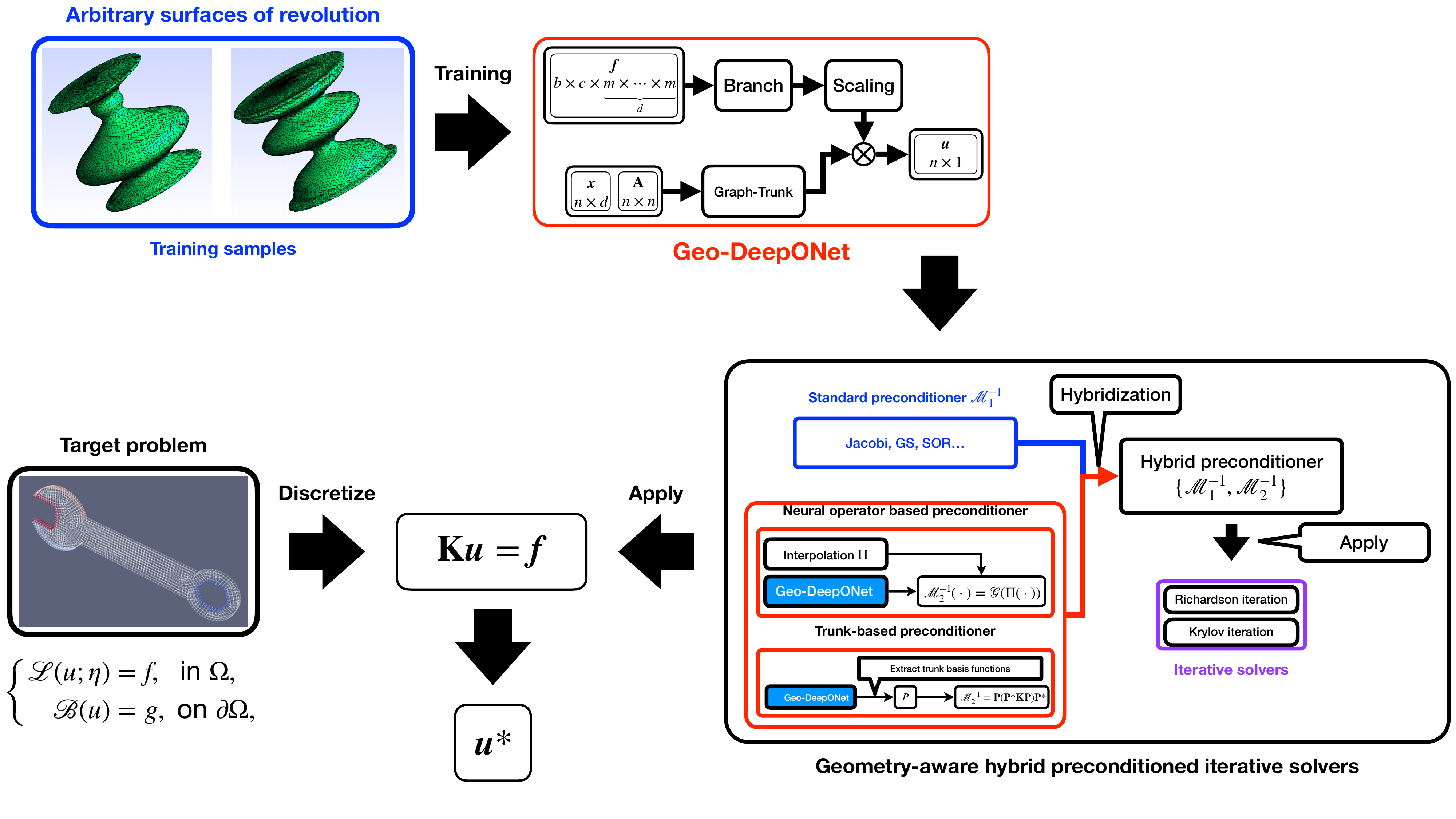}
	\caption{A sketch of the geometry-aware hybrid preconditioned iterative solvers.
		The Geo-DeepONet is trained on training samples consisting of arbitrary generated surfaces of revolution.
		Geo-DeepONet, which encodes various revolution-based geometries, effectively generates preconditioners even if it has not seen the target geometries before.
	}\label{fig:hp}
\end{figure}

In this work we propose hybrid iterative solvers with neural preconditioners to solve parametric PDEs on unstructured domains that were not encountered during the training of the neural operator.
The key idea of handling unstructured domains is to encode the arbitrary geometries into the NOs.
When the given domain is discretized with the FEM, a connectivity matrix of nodes, also known as adjacency matrix, can be obtained straightforwardly. 
Specifically, we utilize a deep operator network~(DeepONet)~\cite{lu2021learning} architecture, which performs well compared to other NOs, especially in unstructured domains~\cite{lu2022comprehensive}.
We first introduce a modified trunk network within the DeepONet, denoted as a \textit{graph-trunk network}, which takes the coordinates and the corresponding nodal connectivity as input; the specific graph convolution used is Chebyshev spectral graph convolutional networks~\cite{defferrard2016convolutional}.
Secondly, we employ the convolutional neural network~(CNN), which performs effectively in 2D or 3D structured grids~\cite{lu2021learning,lu2022comprehensive,raonic2024convolutional} as a branch network and introduce the canonical projection of features from the unstructured grid onto the structured grid.
Specifically, this projection is constructed by interpolating the nodal features of the unstructured mesh onto a uniformly discretized Cartesian grid, ensuring that the geometric information is consistently transferred while preserving the resolution needed for the convolution layers.
Finally, inspired by the squeeze-and-excitation network~\cite{hu2018squeeze}, we design a \textit{scaling network} that acts on the output of the branch network.
This network produces a scaled output that acts like an attention mechanism~\cite{vaswani2017attention} for the given input functions applied to the corresponding unstructured domain.
By combining all subnetworks, we construct a geometry-aware DeepONet~(Geo-DeepONet) that outperforms the plain DeepONet in unstructured domains.
A sketch of the methodology is shown in~\Cref{fig:hp}.
The superior performance of the geometry-aware hybrid preconditioned iterative solvers is verified by various numerical experiments.

This paper is organized as follows: In~\Cref{sec:problem}, we describe the model problem and review the finite element method.
In~\Cref{sec:methodolgy}, we introduce the hybrid preconditioning framework that combines the pre-trained NO with iterative solvers, and propose the Geo-DeepONet.
Finally, in~\Cref{sec:results}, we demonstrate the numerical performance of the proposed geometry-aware hybrid preconditioned iterative solvers.
Conclusions and future work are discussed in~\Cref{sec:conclusion}.

\section{Model problem}\label{sec:problem}
Let $\Omega \subset \R^{d}$, $d \in \{2, 3\}$ be a bounded domain, which is not necessarily convex or simply connected.
Let $\H \subset \R^{P}$ be the parameter space, a closed and bounded subset of $\R^{P}$, where $P \ge 1$ denotes the number of parameters.
For any integer $m \ge 1$ and model parameter $\eta \in \H$, we aim to solve the following parameterized partial differential equation with certain boundary conditions:
\begin{equation}
	\label{eqn:pde}
	\begin{split}
		\A (u; \eta)    & = f \text{ in } \Omega,                                                                            \\
		\B^{k}(u; \eta) & = g^{k} \text{ on } \Gamma^{k} \subset \partial \Omega, \text{ for } k = 1, 2, \ldots, n_{\Gamma},
	\end{split}
\end{equation}
where $\A(\argdot;\eta)$ and ${\{\B^{k}(\argdot;\eta)\}}_{k=1}^{n_{\Gamma}}$ denote a partial differential operator and a set of boundary condition operators in $u$ parameterized by $\eta$, respectively.
Note that we only consider the problem where well-posedness is ensured.
Let $V \subset L^{2}(\Omega)$ be a certain Hilbert space consisting of functions on $\Omega$.
The problem~\eqref{eqn:pde} admits the following weak formulation: Given $\eta \in \H$, find $u \in V$, such that
\begin{equation}
	\label{eqn:weak}
	a(u, v;\eta) = (f, v) \quad \forall v \in V,
\end{equation}
where $a(\argdot, \argdot; \eta)$ and $(\argdot, \argdot)$ denote a bilinear form obtained from $\A (\argdot;\eta)$ and an inner product defined on $L^{2}(\Omega)$, respectively.

In order to solve~\eqref{eqn:weak} using the finite element method, the domain $\Omega$ is discretized into a quasi-uniform triangulation $\T_{h}$ and $V_{h}$ is a finite element space defined on $\T_{h}$.
Note that $h$ is a spatial mesh size.
The finite element solution $u_{h} \in V_{h}$ of~\eqref{eqn:weak} satisfies
\begin{equation}
	\label{eqn:fem}
	a(u_{h}, v;\eta) = (f, v) \quad \forall v \in V_{h}.
\end{equation}

\begin{remark}
	To guarantee convergence to a continuous solution, we need some assumptions on~\eqref{eqn:fem}, such as coercivity, elliptic regularity, and interpolation estimates~\cite{brenner2008mathematical}.
	Since our goal is to reduce the computational cost for solving the linear system induced by~\eqref{eqn:fem}, we only consider the problem that the finite element solution converges to the continuous solution.
\end{remark}

The solution of~\eqref{eqn:fem} can be found by solving the following system of linear equations:
\begin{equation}
	\Km \uv = \fv,
	\label{eqn:linear}
\end{equation}
for the nodal coefficients $\uv \in \R^{n}$.
The stiffness matrix $\Km \in \R^{n \times n}$ and load vector $\fv$ depend affinely on the parameters $\etav$ and their components are computed as follows:
\begin{equation*}
	\Km_{ij} = a(\phi_{i}, \phi_{j}; \eta), \quad \fv_{i} = (f, \phi_{i}), \quad 1 \leq i,j \leq n,
\end{equation*}
where ${\{\phi_{i}\}}_{i=1}^{n}$ denotes a set of nodal basis functions defined in $V_{h}$.

\section{Methodology}\label{sec:methodolgy}
Traditionally, in order to solve the system of linear equations~\eqref{eqn:linear}, various iterative solvers~(Relaxation methods~\cite{saad2003iterative} and Krylov methods~\cite{saad1986gmres,van1992bi}) are typically employed with a certain preconditioner.
However, defining the preconditioner is often challenging when the domain has a non-smooth boundary or includes non-convex regions.
Hybrid preconditioned iterative solvers~\cite{kopanivcakova2024deeponet,lee2025fast,zhang2024blending} address some of these challenges by constructing preconditioners based on pre-trained neural operators, but performance on arbitrary unstructured domains is still an open problem.
In this section, we briefly introduce the basic concepts behind hybrid preconditioned iterative solvers and propose a novel DeepONet-based neural operator that can work for arbitrary unstructured domains.

\subsection{Hybrid preconditioned iterative solvers}

\begin{figure}
	\centering
	\includegraphics[width=\linewidth, trim={3cm, 0cm, 3cm, 0cm}, clip]{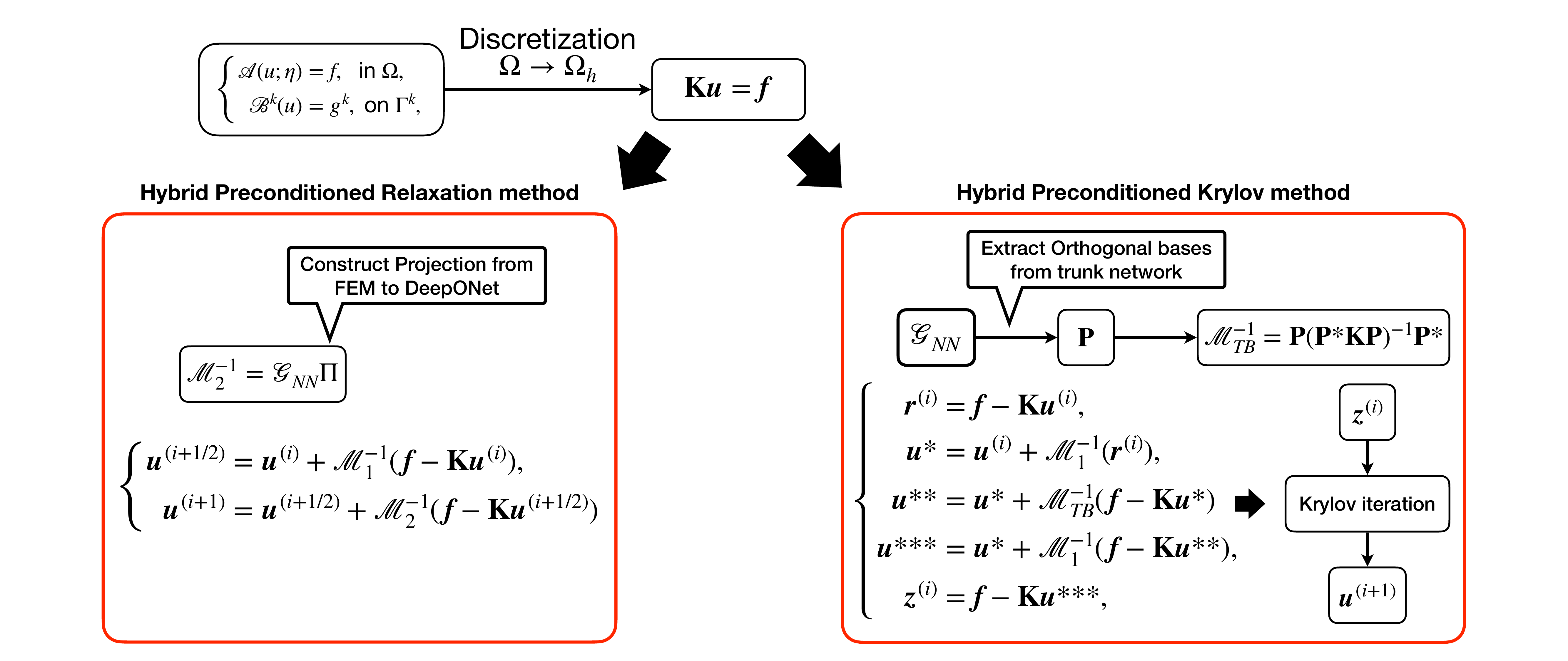}
	\caption{The description of the hybrid preconditioning strategy.
		The hybrid preconditioned relaxation method performs iterations by combining neural operators and classical methods in a certain ratio, leveraging the strengths of each approach to achieve the accurate solution in an efficient way.
        Here $\M_{1}^{-1}$ and $\M_{2}^{-1}$ denote the $n_{r}$ iterations of the relaxation method such as Jacobi or SOR and the prediction of pre-trained DeepONet, respectively.
		On the other hand, in the hybrid preconditioned Krylov method, the prolongation operator $\Pm$ is constructed from the output of the trunk network of the pre-trained DeepONet to define the linear preconditioner $\M_{\text{TB}}^{-1}$.}\label{fig:hints}
\end{figure}

We first describe how to define the hybrid preconditioner for relaxation methods using neural operators.
The standard relaxation method can be formulated as a preconditioned Richardson iteration~\cite{saad1986gmres}.
The $i$-th iterate $\uv^{(i)}$ of the standard relaxation method is given by
\begin{equation*}
	\left\{
	\begin{aligned}
		\rv^{(i)}   & = \fv - \Km\uv^{(i)},            \\
		\uv^{(i+1)} & = \uv^{(i)} + \M^{-1} \rv^{(i)},
	\end{aligned}
	\right.
\end{equation*}
where $\M^{-1}$ denotes the preconditioner of the Richardson iteration.
By choosing $\M^{-1}$ as the diagonal part of $\Km$, we obtain the Jacobi method, while selecting the lower triangular part leads to the Gauss-Seidel~(GS) method.
Utilizing the multiplicative subspace correction framework~\cite{xu1992iterative}, the hybrid preconditioned Richardson iteration~\cite{zhang2024blending}, also known as HINTS, can be written as
\begin{equation}
	\label{eqn:hybrid}
	\left\{
	\begin{aligned}
		\rv^{(i)}     & = \fv - \Km\uv^{(i)},                         \\
		\uv^{(i+1/2)} & = \uv^{(i)} + \M_{1}^{-1}(\rv^{(i)}),         \\
		\rv^{(i+1/2)} & = \fv - \Km\uv^{(i+1/2)},                     \\
		\uv^{(i+1)}   & = \uv^{(i+1/2)} + \M_{2}^{-1}(\rv^{(i+1/2)}), \\
	\end{aligned}
	\right.
\end{equation}
where $\M_{1}^{-1}$ and $\M_{2}^{-1}$ denote the $n_{r}$ iterations of the relaxation method and the prediction of the pre-trained neural operator, respectively.
Further practical details about the second preconditioner $\M_{2}^{-1}$ are described in~\Cref{sec:hints}.

Krylov methods require linear preconditioners, so it is not suitable to directly use the neural operator to precondition them.
To overcome this challenge, one of the simplest approaches is combining the HINTS framework with flexible Krylov methods~\cite{simoncini2002flexible}, which allows the preconditioner to be a nonlinear operator.
However, to tackle this challenge directly, the trunk basis~(TB) approach~\cite{kopanivcakova2024deeponet} was proposed, which extracts the basis functions from the output of the trunk network of the pre-trained DeepONet and uses them to design a linear preconditioner.
We first introduce the TB-based prolongation operator $\Pm \in \R^{n \times k}$, where $k$ represents the number of basis functions used to construct the preconditioner.
The $(i,j)$-th component of the representation matrix of $\Pm$ is given by
\begin{equation*}
	{[\Pm]}_{ij} = T_{j}(\xv_{i}),
\end{equation*}
where $T_{j}(\xv_{i})$ is the $j$-th component of the output of the trunk network evaluated at the coordinate $\xv_{j} \in \Omega$.
The restriction operator $\mathbf{R} \in \R^{k \times n}$ is naturally defined as the adjoint operator of $\mathbf{P}$, i.e., $\mathbf{R}:=\mathbf{P}^{\ast}$.
The TB-based preconditioner $\M_{\text{TB}}^{-1}$ is defined as
\begin{equation}
	\label{eqn:linearized_preconditioner}
	\M_{\text{TB}}^{-1} := \Pm{(\underbrace{\Rm\Km\Pm}_{\Km_{c}})}^{-1}\Rm.
\end{equation}
Here, the operator $\Km_{c}^{-1}$ represents the coarse-level problem, which is commonly used in multigrid methods~\cite{briggs2000multigrid}.
It is obvious that the preconditioner defined in~\eqref{eqn:linearized_preconditioner} can be used as the preconditioner for standard Krylov methods.
Finally, given the residual $\rv^{(i)} = \fv - \Km \uv^{(i)}$ in a Krylov iteration, the hybrid preconditioned residual $\zv^{(i)}$ is defined as
\begin{equation*}
	\left\{
	\begin{aligned}
		\uv^{\ast}         & = \uv^{(i)} + \M_{1}^{-1}(\rv^{(i)}),                    \\
		\uv^{\ast\ast}     & = \uv^{\ast} + \M_{\text{TB}}^{-1}(\fv - \Km\uv^{\ast}), \\
		\uv^{\ast\ast\ast} & = \uv^{\ast\ast} + \M_{1}^{-1}(\fv - \Km\uv^{\ast\ast}), \\
		\zv^{(i)}          & = \fv - \Km\uv^{\ast\ast\ast},
	\end{aligned}
	\right.
\end{equation*}
where $\M_{1}^{-1}$ denotes the $n_{r}$ iterations of the relaxation method.
Further practical details regarding the prolongation operator $\Pm$ can be found in~\Cref{sec:tb}.
A schematic view of hybrid preconditioned iterative solvers is depicted in~\Cref{fig:hints}.

\subsection{Geometry-aware deep operator network}
The next challenge is to design and train the neural operator in arbitrary unstructured domains.
In this work, we propose a geometry-aware deep operator network~(Geo-DeepONet) that encodes the geometric information of the arbitrary unstructured domain.

We first briefly describe the mathematical formulation of the standard deep operator network~(DeepONet)~\cite{lu2021learning}.
Let $\X, \Y$ be infinite-dimensional Banach spaces.
Let $\G \colon \X \to \Y$ be an operator to approximate using DeepONet.
Let us call its approximation by $\G_{\text{NN}}$.
The output of DeepONet is defined as the inner product of two distinct neural networks, called the branch network and the trunk network.
The branch network $B \colon \R^{m} \to \R^{p}$ is a vector-valued neural network of dimension $p$ that takes a discretized function $\fv_{m} \in \X_{m}$ as input.
Here, $\X_{m}$ is a $m$-dimensional space satisfying
\begin{align*}
	\forall f \in \X, \quad \exists \fv_{m} \in \X_{m} \text{ such that } \fv_{m} \to f \text{ as } m \to \infty.
\end{align*}
In addition, the trunk network $T \colon \Omega \subset \R^{d} \to \R^{p}$ is another vector-valued neural network of dimension $p$ that takes coordinates in the domain $\Omega$ as input.
The output of DeepONet $\G_{\text{NN}}$ is given by
\begin{equation*}
	\G_{\text{NN}}(\fv)(\xv) := \langle B(\fv), T(\xv) \rangle, \quad \forall \fv \in \X_{m}, \xv \in \Omega,
\end{equation*}
where the symbol $\langle \argdot, \argdot \rangle$ denotes the inner product.
The universal approximation theorem for operators, as proven in~\cite{chen1995uat,lu2021learning}, guarantees that DeepONet can sufficiently approximate the target operator in $L^{2}$ norm.

\begin{figure}
	\centering
	\subfloat[$\bar{\Omega}_{h}$~(Structured mesh)]{
		\includegraphics[width=0.225\linewidth]{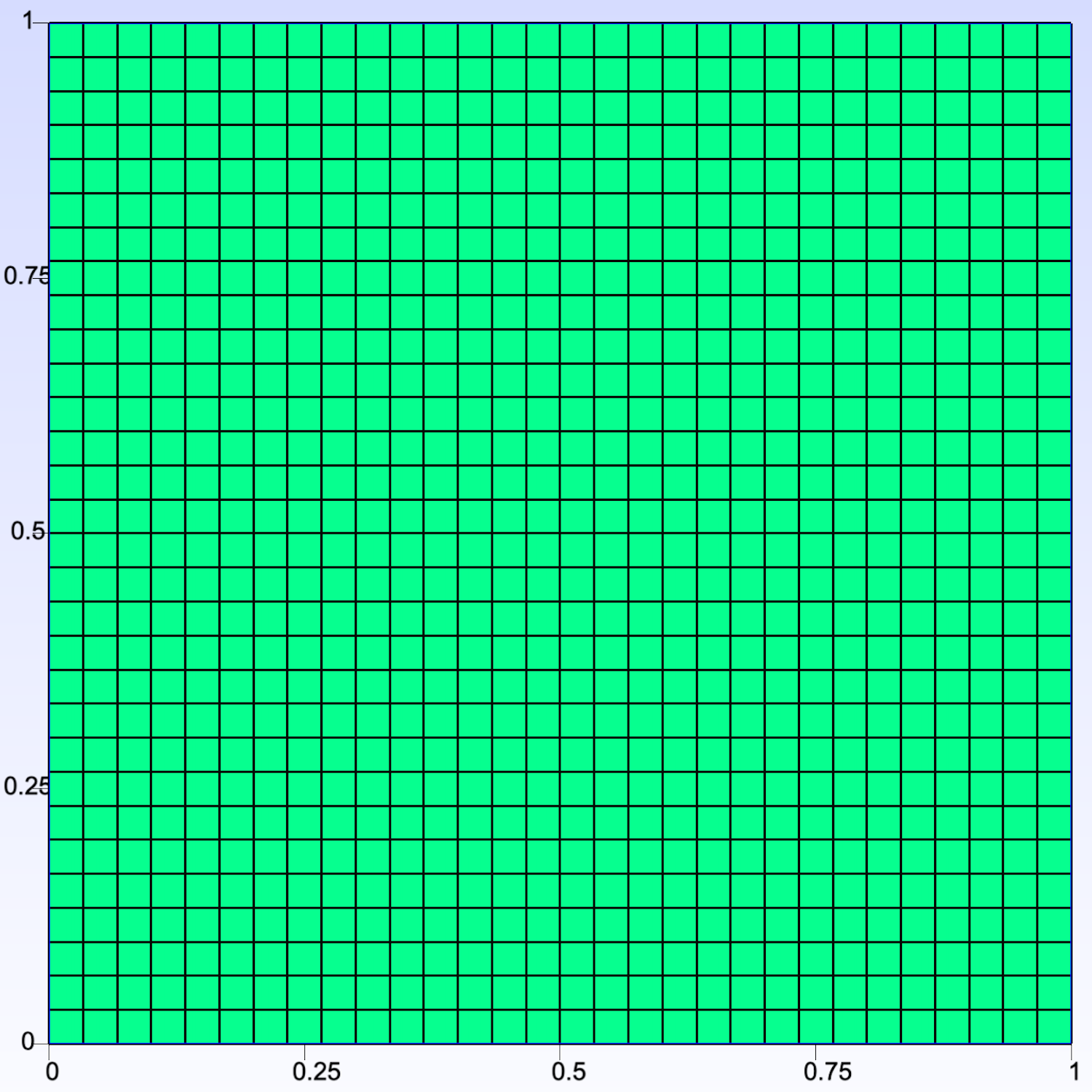}
		\includegraphics[width=0.225\linewidth]{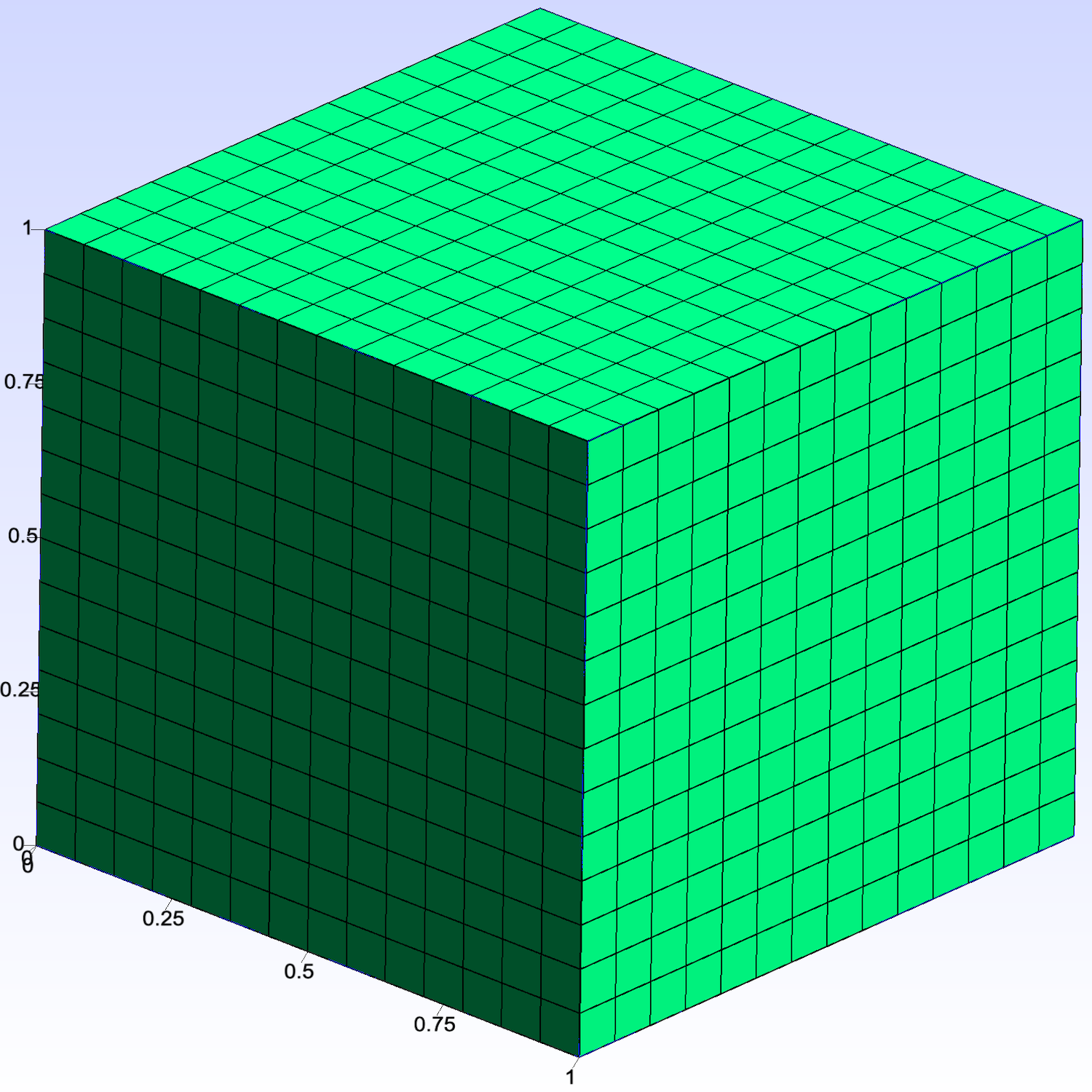}
	}
	\subfloat[$\T_{h}$~(Unstructured mesh)]{
		\includegraphics[width=0.225\linewidth]{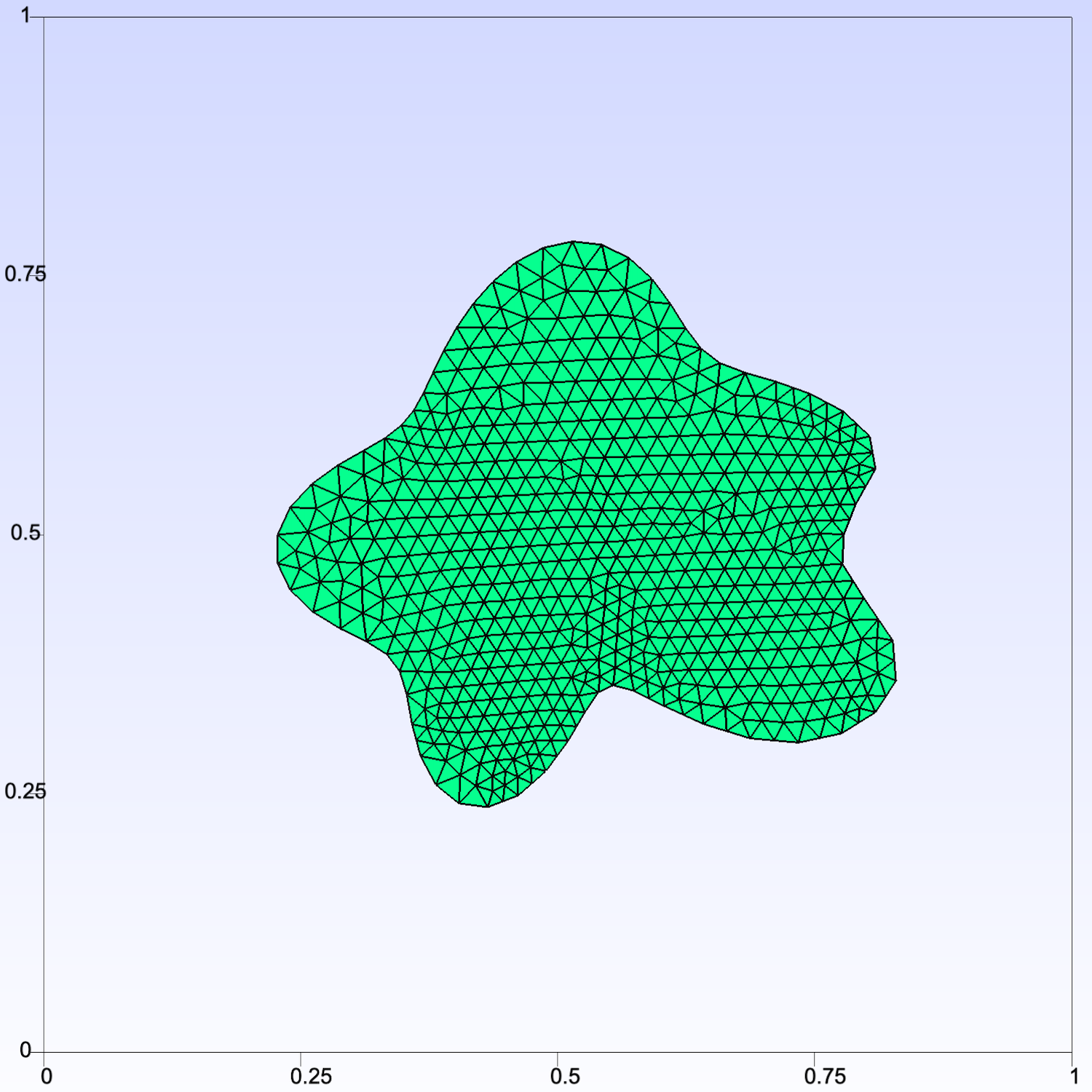}
		\includegraphics[width=0.225\linewidth]{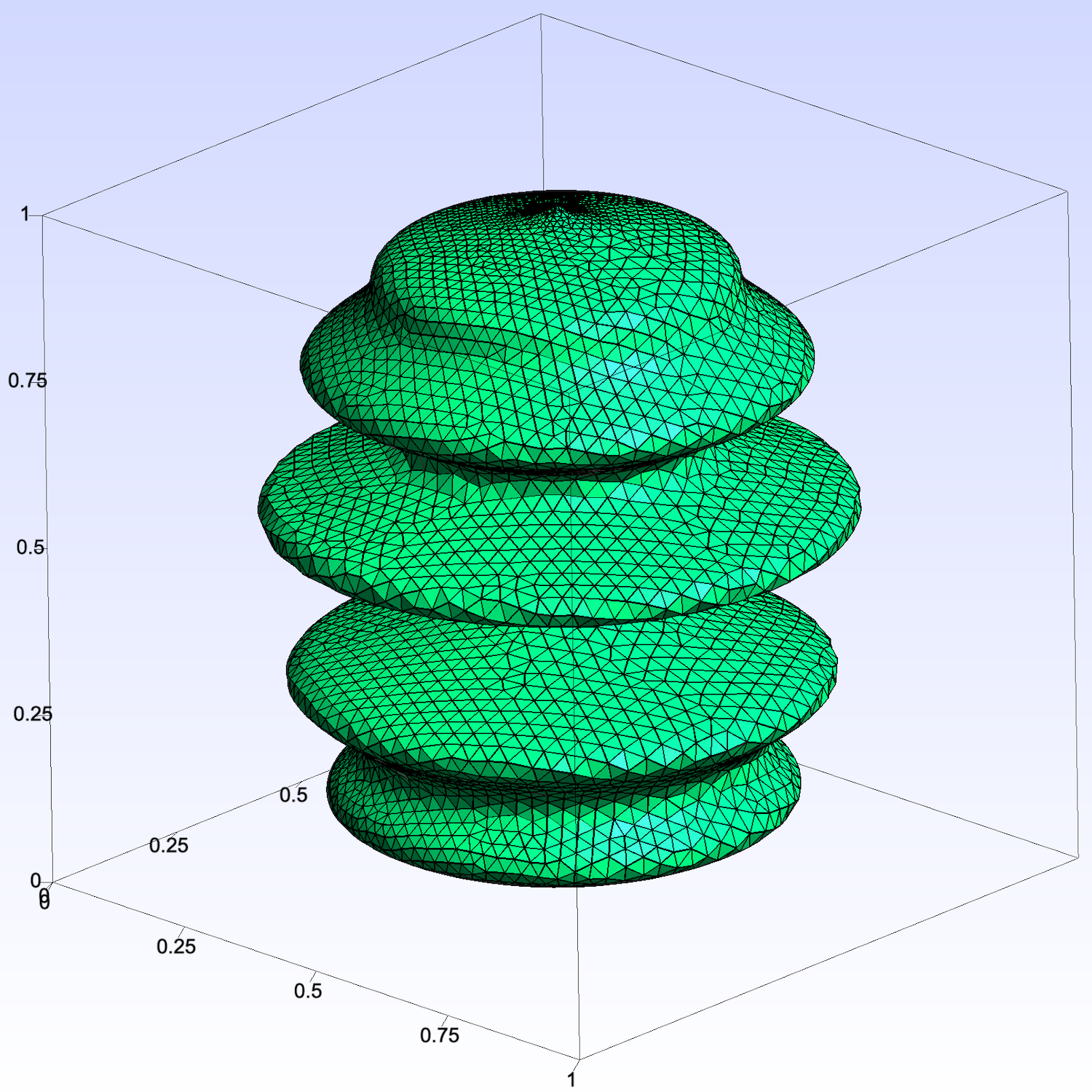}
	}
	\caption{Schematic view of a 2D mesh (left) and a 3D mesh (right) discretization for $\bar{\Omega}_{h}$~(a) and $\T_{h}$~(b), respectively. The maximum size of mesh is set to $h=1/30$ for 2D and $h=1/14$ for 3D, respectively.
	}\label{fig:masked_domain}
\end{figure}

Next, we introduce an essential network, called the graph-trunk network $\tilde{T}$, to encode the unstructured geometries.
Without loss of generality, assume that $\Omega \subset \bar{\Omega} :={[0, 1]}^{d}$.
In addition, we assume that $\bar{\Omega}$ is uniformly discretized with the mesh size $h=1/m$.
We denote this discretized domain as $\bar{\Omega}_{h}$.
Recall that we solve problem~\eqref{eqn:fem} with quasi-uniform triangulation $\T_{h}$.
A schematic view of $\bar{\Omega}_{h}$ and $\T_{h}$ for 2D/3D domains is presented in~\Cref{fig:masked_domain}.

Given the triangulation $\T_{h}$, the node-connectivity matrix $\Am$ can be easily extracted, which is known as the adjacency matrix.
Specifically, the $(i,j)$-th component of $\Am$ is 1 if the $i$-th and $j$-th nodes are connected, and 0 otherwise.
The graph-trunk network $\tilde{T}$ consists of two parts: first, a Chebyshev spectral graph convolutional network~\cite{defferrard2016convolutional}, which has demonstrated success in various graph-related tasks, creates a geometry-feature vector $\yv$ of the domain geometry using the coordinates $\xv$ and the adjacency matrix $\Am$; second, a fully connected neural network takes the geometry-feature vector $\yv$ and produces the basis functions.

The branch network uses a convolutional neural network~(CNN), which has shown great performance on 2D or 3D structured grids~\cite{lu2021learning,lu2022comprehensive,raonic2024convolutional}.
However, the discretized input function $\fv_{m}$ is defined in $\T_{h} \not\subset\bar{\Omega}_{h}$, which requires the projection $\Pi_{h}$ from $\T_{h}$ to $\bar{\Omega}_{h}$.
Further details of this projection can be found in~\Cref{sec:hints}.

In addition, to enhance the performance of the CNN in the branch network, we introduce another neural network called the scaling network $S$, inspired by the squeeze-and-excitation network~\cite{hu2018squeeze}. In the scaling network, the output of the branch network is split into two paths. The first path passes through two fully connected layers and the softmax function, while the second path passes through a single fully connected layer. Finally, the outputs from both paths are element-wise multiplied to produce the scaled output. This scaling technique has been used in numerous studies to significantly improve the performance of CNNs across various fields~\cite{hu2018squeeze,tan2019mnas,tan2019efficient}.

\begin{figure}
	\centering
	\subfloat[Branch network]{
		\includegraphics[width=0.45\linewidth, trim={7cm 10cm 7cm 10cm}, clip]{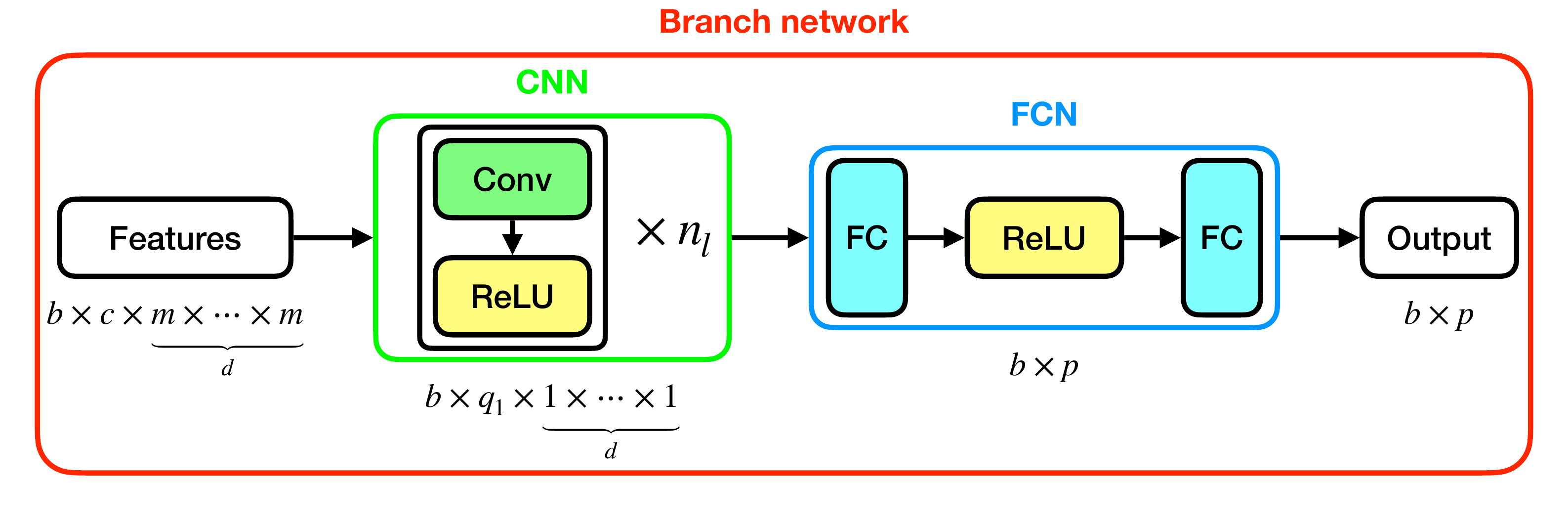}
	}
	\subfloat[Scaling network]{
		\includegraphics[width=0.45\linewidth, trim={7cm 10cm 7cm 9cm}, clip]{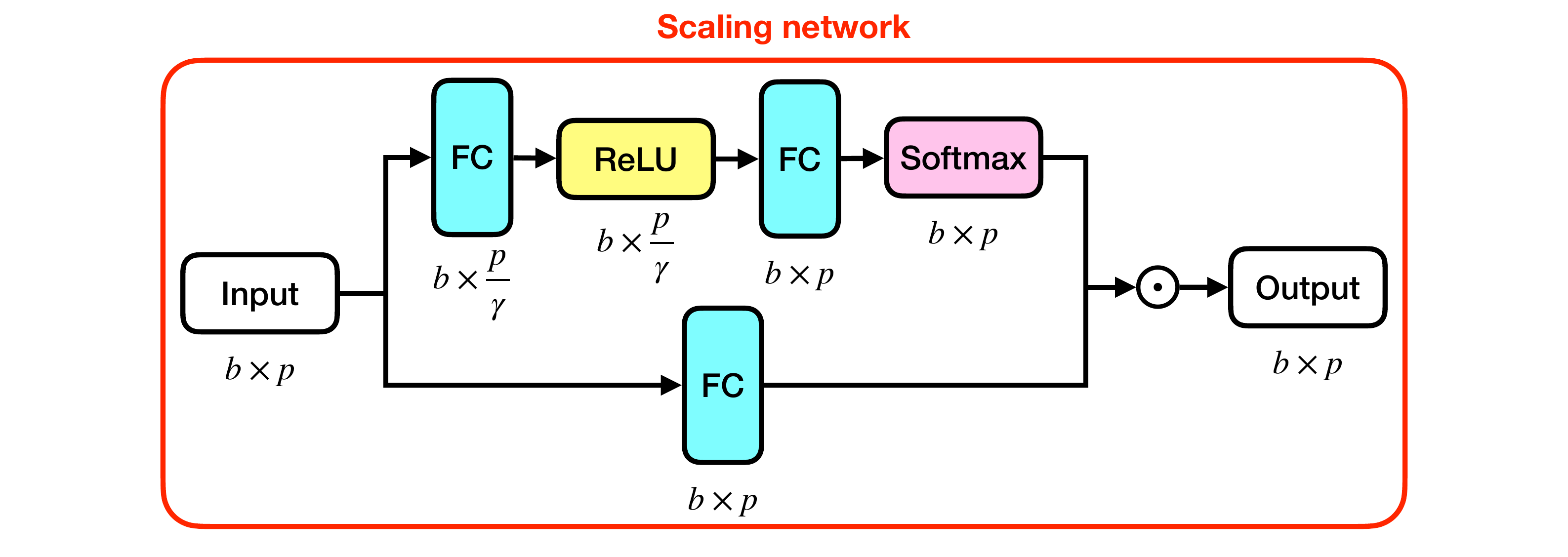}
	}
	\\
	\subfloat[Graph-trunk network]{
		\includegraphics[width=0.45\linewidth, trim={10cm 8cm 10cm 7cm}, clip]{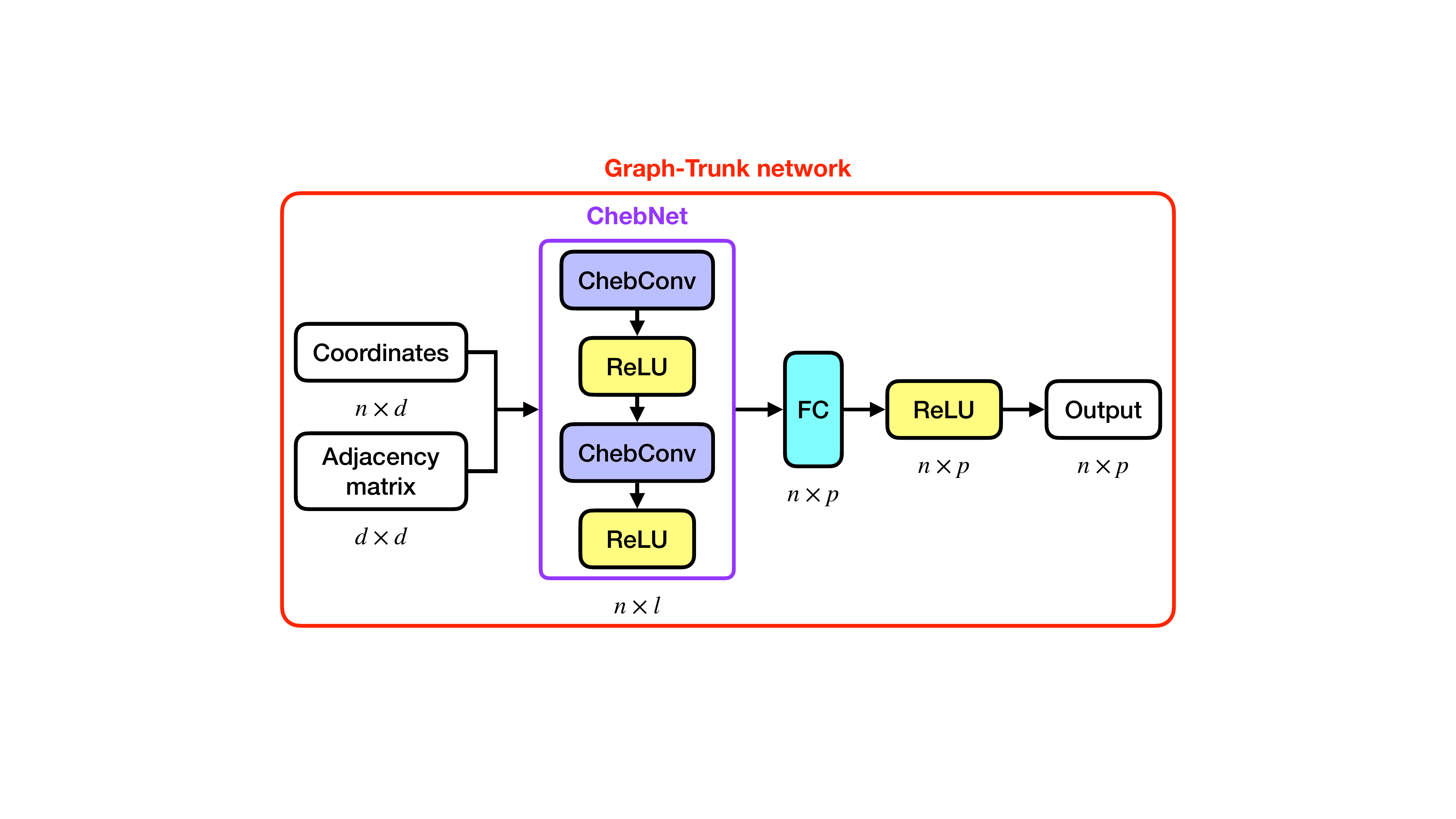}
	}
	\subfloat[Geo-DeepONet]{
		\includegraphics[width=0.45\linewidth, trim={15cm 3cm 15cm 3cm}, clip]{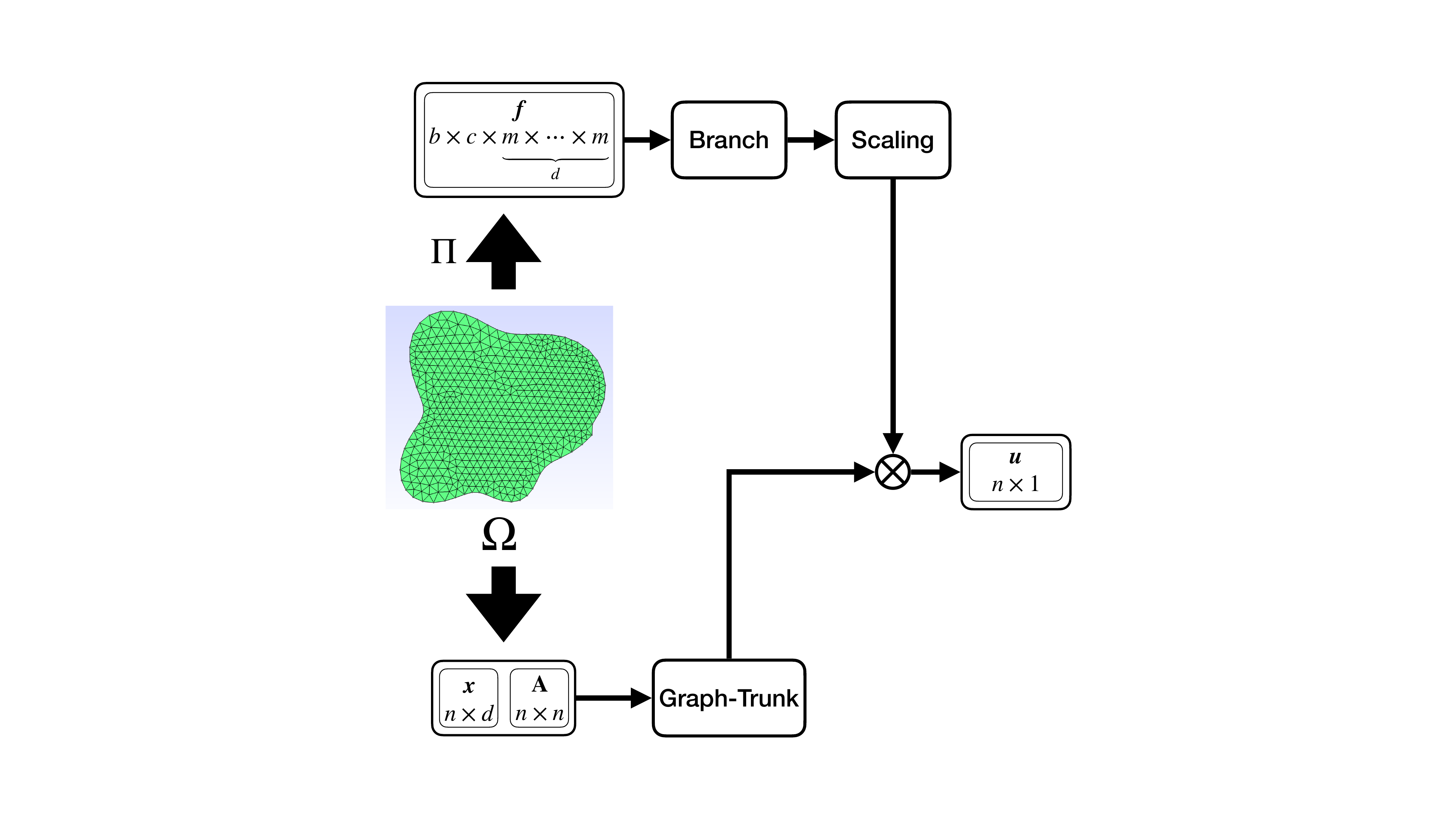}
	}
	\caption{An illustration of Geo-DeepONet. For a given unstructured domain $\Omega$, the projected features $\Pi \fv$, the coordinates $\xv$, and the adjacency matrix $\Am$ are constructed. The prediction of Geo-DeepONet is defined as the inner product of the output of the scaling network and the graph-trunk network.}\label{fig:geonet}
\end{figure}

Finally, the Geo-DeepONet $\G_{\text{geo}}$ is defined as
\begin{equation*}
	\G_{\text{geo}}(\fv, \xv, \Am) := \langle S \circ B(\Pi_{h} \fv), \tilde{T}(\xv, \Am) \rangle.
\end{equation*}
The overall structure of the proposed Geo-DeepONet is depicted in~\Cref{fig:geonet}.

\subsubsection{Training Geo-DeepONet}
To train the proposed Geo-DeepONet, we construct the dataset $\D$ of $N_{s}$ samples given by
\begin{equation*}
	\D = {\{ (\Omega_{j}, \fv_{j}, \xv_{j}, \Am_{j}, \uv_{j}) \}}_{j=1}^{N_{s}}.
\end{equation*}
The coordinate $\xv_{j}$ and the adjacency matrix $\Am_{j}$ are determined by discretization of $\Omega_{j}$.
The symbols $\fv_{j}$ and $\uv_{j}$ denote the projected input function and the reference solution computed in $\Omega_{j}$, respectively.
The training process of the Geo-DeepONet is conducted by minimizing the point-wise relative squared error~\cite{kahana2023geometry,lu2021learning,lu2022comprehensive,zhang2024blending} between the prediction of network and the reference solution:
\begin{equation*}
	\argmin_{\thetav} \frac{1}{N_{s}}\sum_{j=1}^{N_{s}} \frac{\lvert \G_{\text{geo}} (\fv_{j}, \xv_{j}, \Am_{j}) - \uv_{j} \rvert^{2}}{\lvert \uv_{j} \rvert^{2}},
\end{equation*}
where $\thetav$ denotes all parameters in the Geo-DeepONet.

The solution of a parametric PDE is highly dependent on the boundary conditions.
Basically, to train the DeepONet to solve the problem~\eqref{eqn:pde}, it is required to sample the right-hand side function $f$ and the boundary functions ${\{g_{k}\}}_{k=1}^{n_{\Gamma}}$.
As the number of boundary conditions increases, the number of functions to be sampled also increases.
This not only requires more computational resources to prepare the training dataset, but also requires the DeepONet to add additional branch networks.
However, when used in the hybrid preconditioning framework described in~\eqref{eqn:hybrid}, the DeepONet does not require boundary functions for the training process. Since DeepONet acts on the residual $\rv^{(i+1/2)}$, it is enough for DeepONet to be trained on problems with homogeneous boundary conditions~\cite{lee2026two}.
This property allows us to bypass the need to generate boundary functions for training data, reducing memory requirement, and simplifying the structure of DeepONet.
The mathematical details of this property are described in~\Cref{sec:homogeneous}.

\section{Numerical results}\label{sec:results}
In this section we report the numerical results of the proposed geometry-aware hybrid preconditioned iterative solvers and compare the performance for different PDEs in unstructured domains.
These problems have been selected because
they are established as standard benchmark problems.

\subsection{Benchmark problems}
We describe the benchmark problems, formulated in two or three dimensions, and demonstrate the performance of the proposed hybrid preconditioned iterative solvers.

\subsubsection{Poisson equation with a variable coefficient}\label{sec:poisson}
We first consider a Poisson equation in an unstructured domain $\Omega \subset \R^{d}, d=2,3$, which is given as
\begin{equation*}
	\begin{aligned}
		- \nabla \cdot (\kappa (\xv) \nabla u(\xv) ) & = f(\xv), \qquad  \  \forall \xv \in \Omega,           \\
		u(\xv)                                       & = 0, \qquad \qquad \quad  \text{on} \ \partial \Omega,
	\end{aligned}
\end{equation*}
where $u, f$ denote the solution and the forcing term, respectively.
To create the training samples, we sample the diffusion coefficient~$\kappa \geq 0.3$ using Gaussian random fields~(GRFs) with mean $\E[\kappa (\xv)]=1.0$ and  the covariance given as
\begin{align}
	\text{Cov}(\kappa(\xv), \kappa( \xv^{\prime})) = \sigma_{g}^{2} \exp \left(- \frac{\lVert \xv - \xv^{\prime} \rVert^{2}}{2 \ell^{2}} \right), \quad \xv, \xv^{\prime} \in \Omega.
	\label{eqn:cov}
\end{align}
The parameters~$\sigma_{g}$ and $\ell$ are chosen as $\sigma_{g} = 0.3$ and $\ell = 0.1$.
The forcing term~$f$ is also sampled using GRFs, but with zero mean and the covariance given as in~\eqref{eqn:cov}, but with
parameters $\sigma_{g} = 1.0$ and $\ell = 0.1$.

\subsubsection{Linear elasticity equation}
We next consider a linear elasticity equation in an unstructured domain $\Omega \subset \R^{d}, d=2,3$, which is given by
\begin{equation*}
	\begin{aligned}
		-\nabla \cdot \sigma\left(\uv\right) & = \fv(\xv)                                                                                                      & \text{ in } \Omega,                        \\
		\sigma\left(\uv\right)               & = \lambda(\xv) \mathrm{tr}\left(\varepsilon\left(\uv\right)\right)\Id + 2 \mu(\xv) \varepsilon\left(\uv\right),                                              \\
		\varepsilon (\uv)                    & = \frac{1}{2} ( \nabla \uv + {(\nabla \uv)}^{T}),                                                                                                            \\
		\uv                                  & = 0,                                                                                                            & \text{ on } \Gamma,                        \\
		\frac{\partial \uv}{\partial n}      & = 0,                                                                                                            & \text{ on } \partial\Omega\setminus\Gamma,
	\end{aligned}
\end{equation*}
where~$\uv, \fv\in \R^{d}$ denote the displacement and the body force, respectively.
The symbols~$\sigma, \varepsilon$ denote the stress tensor and the linearized strain tensor, respectively.
The material parameters $\lambda$ and $\mu$ are computed by
\begin{align*}
	\lambda (\xv) = \frac{\nu E(\xv)}{(1+\nu)(1-2\nu)}, \quad \varepsilon (\xv) = \frac{E(\xv)}{2(1+\nu)},
\end{align*}
where the Poisson's ratio is set to $\nu=0.3$ and the Young's modulus $E(\xv)$ is sampled using GRFs following the same approach as the diffusion coefficient $\kappa$ in~\Cref{sec:poisson}.
Similarly, the body force $\fv$ is sampled using the same strategy applied to the forcing term~$f$ in \Cref{sec:poisson}.

\subsection{Implementation details}

\begin{table}
	\centering
	\caption{The summary of Geo-DeepONets' architectures.}\label{tab:architecture}
	\footnotesize
	\begin{tabular}{lrr}
		\toprule
		\multirow{2}{*}{\textbf{Problem}} & \multicolumn{2}{c}{\textbf{Branch network}}                       \\ \cline{2-3}
		                                  & \multicolumn{1}{c|}{\textbf{Layers}}             & \textbf{Act.}  \\ \midrule\midrule
		{Poisson 2D}                      & Conv2D[2, 40, 60, 100, 180] + FC[180, 256, 128]  & ReLU           \\ \midrule
		{Poisson 3D}                      & Conv3D[2, 40, 70, 130, 220] + FC[220, 512, 256]  & ReLU           \\ \midrule
		{Elasticity 2D}                   & Conv2D[3, 40, 80, 160, 280] + FC[280, 512, 256]  & ReLU           \\ \midrule
		{Elasticity 3D}                   & Conv3D[4, 40, 80, 240, 480] + FC[480, 1152, 768] & ReLU           \\ \midrule
		\midrule\midrule
		\multirow{2}{*}{\textbf{Problem}} & \multicolumn{2}{c}{\textbf{Scaling network}}                      \\ \cline{2-3}
		                                  & \multicolumn{1}{c|}{  \textbf{Layers} }          & \textbf{Act.}  \\ \midrule
		{Poisson 2D}                      & FC[128, 8, 128] / FC[128, 128]                   & ReLU / Softmax \\ \midrule
		{Poisson 3D}                      & FC[256, 16, 256] / FC[256, 256]                  & ReLU / Softmax \\ \midrule
		{Elasticity 2D}                   & FC[256, 16, 256] / FC[256, 256]                  & ReLU / Softmax \\ \midrule
		{Elasticity 3D}                   & FC[768, 48, 768] / FC[768, 768]                  & ReLU / Softmax \\ \midrule
		\midrule\midrule
		\multirow{2}{*}{\textbf{Problem}} & \multicolumn{2}{c}{\textbf{Graph-trunk network}}                  \\ \cline{2-3}
		                                  & \multicolumn{1}{c|}{  \textbf{Layers} }          & \textbf{Act.}  \\ \midrule
		{Poisson 2D}                      & ChebConv[2, 256, 256] + FC[256, 128]             & ReLU           \\ \midrule
		{Poisson 3D}                      & ChebConv[3, 512, 512] + FC[512, 256]             & ReLU           \\ \midrule
		{Elasticity 2D}                   & ChebConv[2, 256, 256] + FC[256, 128]             & ReLU           \\ \midrule
		{Elasticity 3D}                   & ChebConv[3, 512, 512] + FC[512, 256]             & ReLU           \\
		\bottomrule
	\end{tabular}
\end{table}

In this section, we provide details regarding the hybrid preconditioned iterative solvers for all benchmark problems.
To generate arbitrary 2D/3D domains, an arbitrary smooth curve is sampled from a one-dimensional GRF.\@
Here, we set the mean to $0.2$ and the covariance given as in~\eqref{eqn:cov} with parameters $\sigma_{g}=0.2$ and $\ell=0.1$, ensuring that the values range from $0.1$ to $0.5$.
In the 2D case, a closed curve centered at
$(0.5,0.5)$ was constructed, while in the 3D case, a surface of revolution was generated around the axis centered at $(0.5,0.5,1)$.
After that, all meshes are generated by Gmsh~\cite{geuzaine2009gmsh} with a maximum mesh size $h=1/128$~(2D) or $h=1/32$~(3D).
The generated domains are used to train all networks and their performances are compared in~\Cref{sec:ablation} and~\Cref{sec:convergence}.

We utilize $P^{1}$ conforming finite elements in the benchmark problems and generate training samples using the FEniCSx library~\cite{ScroggsEtal2022}.
Note that the reference solutions are computed using the sparse direct solver MUMPS~\cite{MUMPS:2,MUMPS:1}.
For the 2D case, we use a mesh size of $h=1/30$, while for the 3D case, we set $h=1/14$.
We generate $50,000$ training samples and $10,000$ test samples for all benchmark problems.
The hybrid preconditioned iterative solvers are implemented using the PETSc library~\cite{balay2019petsc} under the petsc4py interface.
We use Jacobi, successive over-relaxation~(SOR)~\cite{young1954iterative}, conjugate gradient~(CG)~\cite{hestenes1952methods}, flexible CG~(FCG)~\cite{notay2000flexible}, generalized minimal residual method~(GMRES)~\cite{saad1986gmres}, and flexible GMRES~(FGMRES)~\cite{saad1993flexible} as the backbone iterative solver.
Geo-DeepONets are implemented using PyTorch~\cite{paszke2019hh} and trained using the AdamW optimizer~\cite{loshchilov2018decoupled}, with a batch size of $100$ and a learning rate of $10^{-4}$.
The training process stops if the average relative $L^{2}$ error of the validation samples falls below $8\%$ or the training loss value does not improve for $100$ consecutive epochs.
Note that the relative $L^{2}$ error at $i$-th epoch is given by
\begin{equation*}
	\Er_{\text{rel}}(u_{\theta}, u_{\text{ref}}) = \frac{\Vert u_{\theta} - u_{\text{ref}}\Vert_{2}}{\Vert u_{\text{ref}}\Vert_{2}},
\end{equation*}
where $u_{\theta}$ and $u_{\text{ref}}$ denote the prediction of the neural operator and the reference solution, respectively.
\Cref{tab:architecture} shows the details of the architecture of Geo-DeepONets for benchmark problems.
In all experiments, the hybrid preconditioned iterative solvers terminate when either of the following criteria is satisfied
\begin{equation*}
	\Vert \rv^{(i)}\Vert_{2} \leq 10^{-12} \quad \text{ or } \quad \frac{\Vert \rv^{(i)}\Vert_{2}}{\Vert \rv^{(0)}\Vert_{2}} \leq 10^{-8}.
\end{equation*}
All experiments were conducted using computational resources and services at the Center for Computation and Visualization, Brown University.
Each computing node is equipped with an AMD EPYC 9554 64-Core Processor~(256GB) and an NVIDIA L40S GPU~(48GB).

\subsection{Ablation study}\label{sec:ablation}

\begin{figure}
	\centering
	\subfloat[$\Omega={(0,1)}^{2}$~(Unit square)]{
	\includegraphics[width=0.9\textwidth, trim={1cm, 2cm, 1cm, 2cm}, clip]{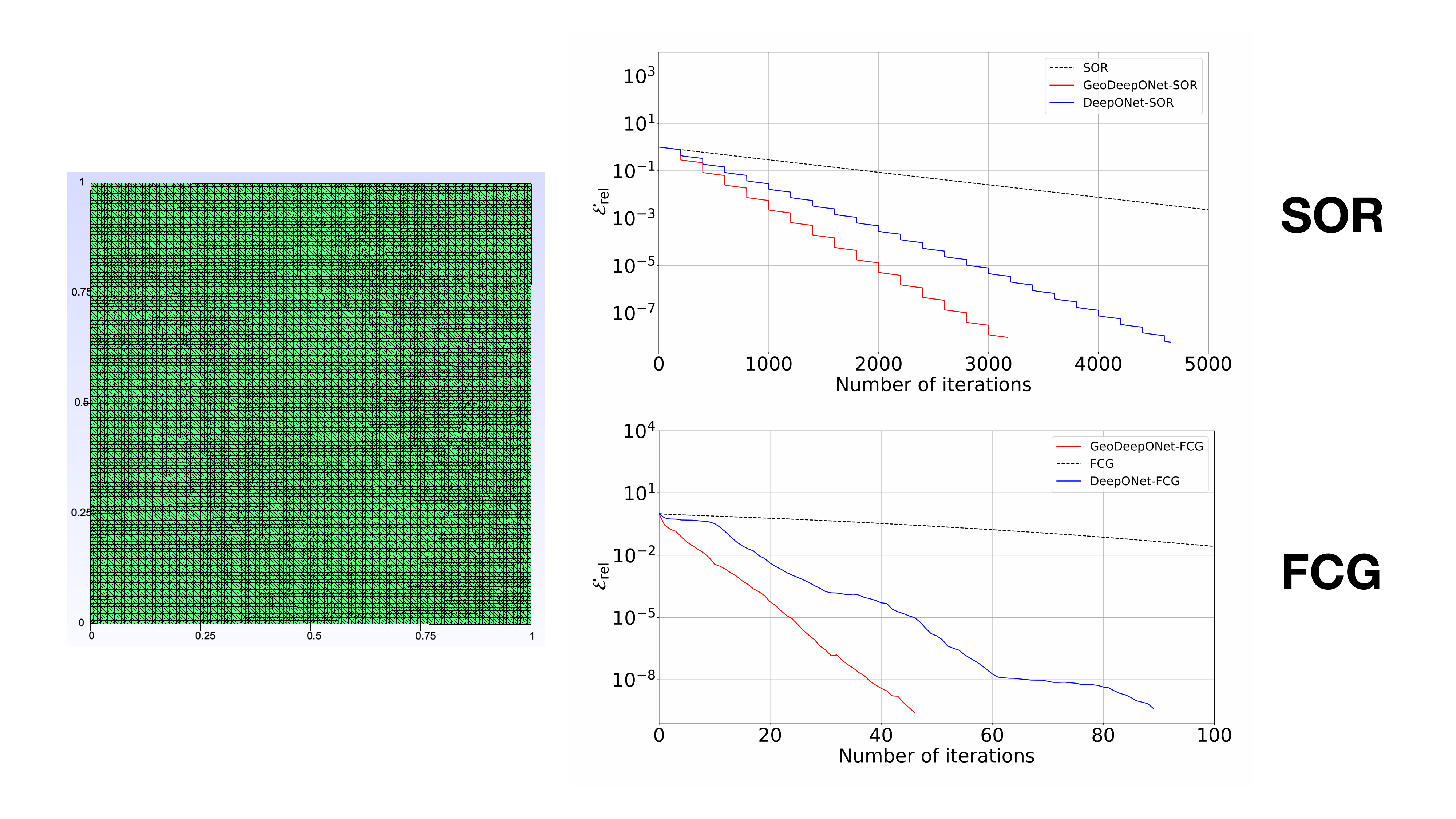}
	}
	\newline
	\subfloat[$\Omega \subset {(0,1)}^{2}$~(Unstructured)]{
		\includegraphics[width=0.9\textwidth, trim={1cm, 2cm, 1cm, 2cm}, clip]{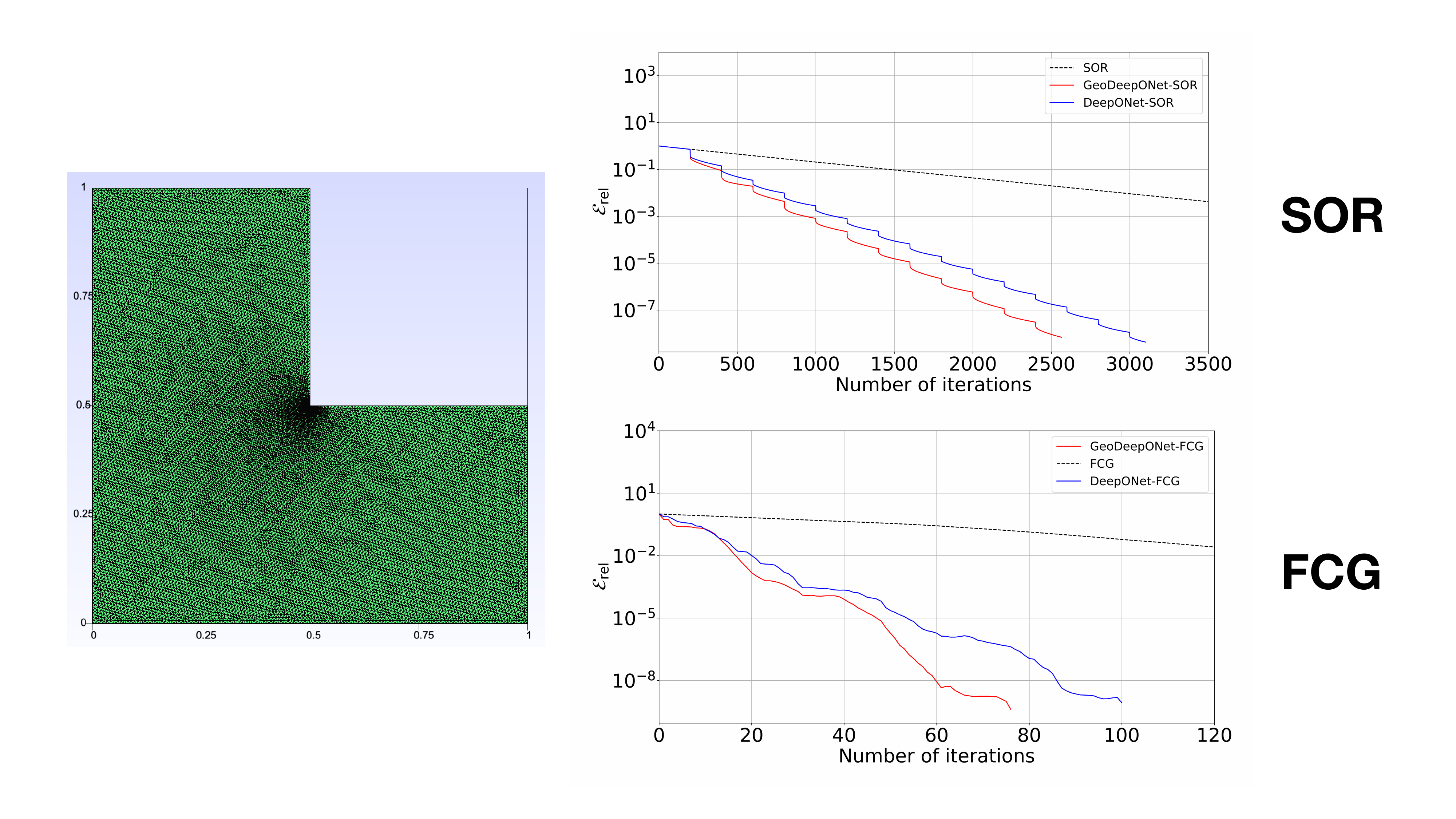}
	}
	\caption{The convergence of SOR and FCG variants for 2D Poisson equation~($\kappa$ is sampled using 2D GRF described in~\Cref{sec:poisson} and $f \equiv 1$) in structured domain~(a) and unstructured domain~(b). DeepONet-\{SOR,FCG\}~(blue) and GeoDeepONet-\{SOR,FCG\}~(red) are created by the hybrid preconditioning framework described in~\eqref{eqn:hybrid}.
		Note that SOR is used for the relaxation step in FCG based solvers.
		All domains are discretized by Gmsh~\cite{geuzaine2009gmsh} with a maximum mesh size $h=1/128$.
		The number of relaxation steps is set to $n_{r}=100$ for SOR and $n_{r}=1$ for FCG.}\label{fig:ablation}
\end{figure}

In this section, experiments for the ablation study were conducted to highlight the effect of the proposed Geo-DeepONet.
We compare the generalization performance of the standard DeepONet and Geo-DeepONet when they are used as the backbone of hybrid preconditioned SOR or FCG.\@
The standard DeepONet employs fully connected layers~(width [2, 256, 256, 256, 128]) for the trunk network, while Geo-DeepONet uses the structure described in~\Cref{tab:architecture}.
Note that the standard DeepONet does not use the scaling network.
Both networks are trained on the same training dataset consisting of arbitrary unstructured domains until they have less than $8\%$ relative $L^{2}$ error.
It is important to point out that the unstructured domain for testing is one that all networks have not seen in the training phase.
\Cref{fig:ablation} illustrates the decay of the relative $L^{2}$ error of the 2D Poisson equation in the structured and unstructured domains.
We can observe that the performance of Geo-DeepONet is much better than that of the standard DeepONet in the structured and unstructured cases, despite being trained under the exact same conditions.
That is, the scaling network and the graph-trunk network effectively encode arbitrary geometries into the neural operator, enhancing the generalizability of Geo-DeepONet over the standard DeepONet in terms of geometric complexity.

\subsection{Performance of the proposed geometry-aware hybrid preconditioned iterative solvers}\label{sec:convergence}

\begin{figure}
	\centering
	\subfloat[Poisson 2D]{
		\includegraphics[width=0.7\textwidth, trim={1cm, 2cm, 1cm, 4cm}, clip]{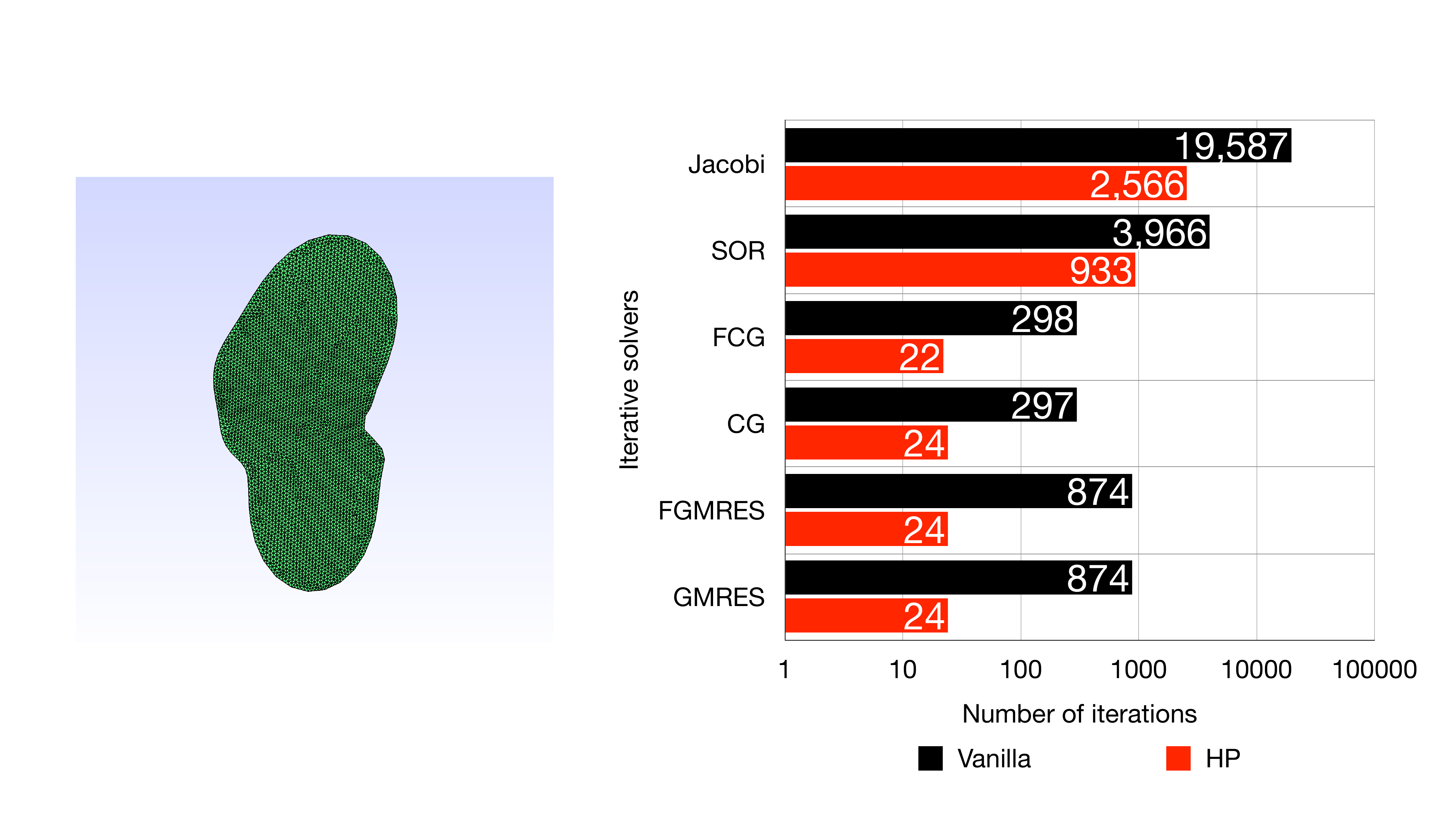}
	}
	\newline
	\subfloat[Poisson 3D]{
		\includegraphics[width=0.7\textwidth, trim={1cm, 2cm, 1cm, 4cm}, clip]{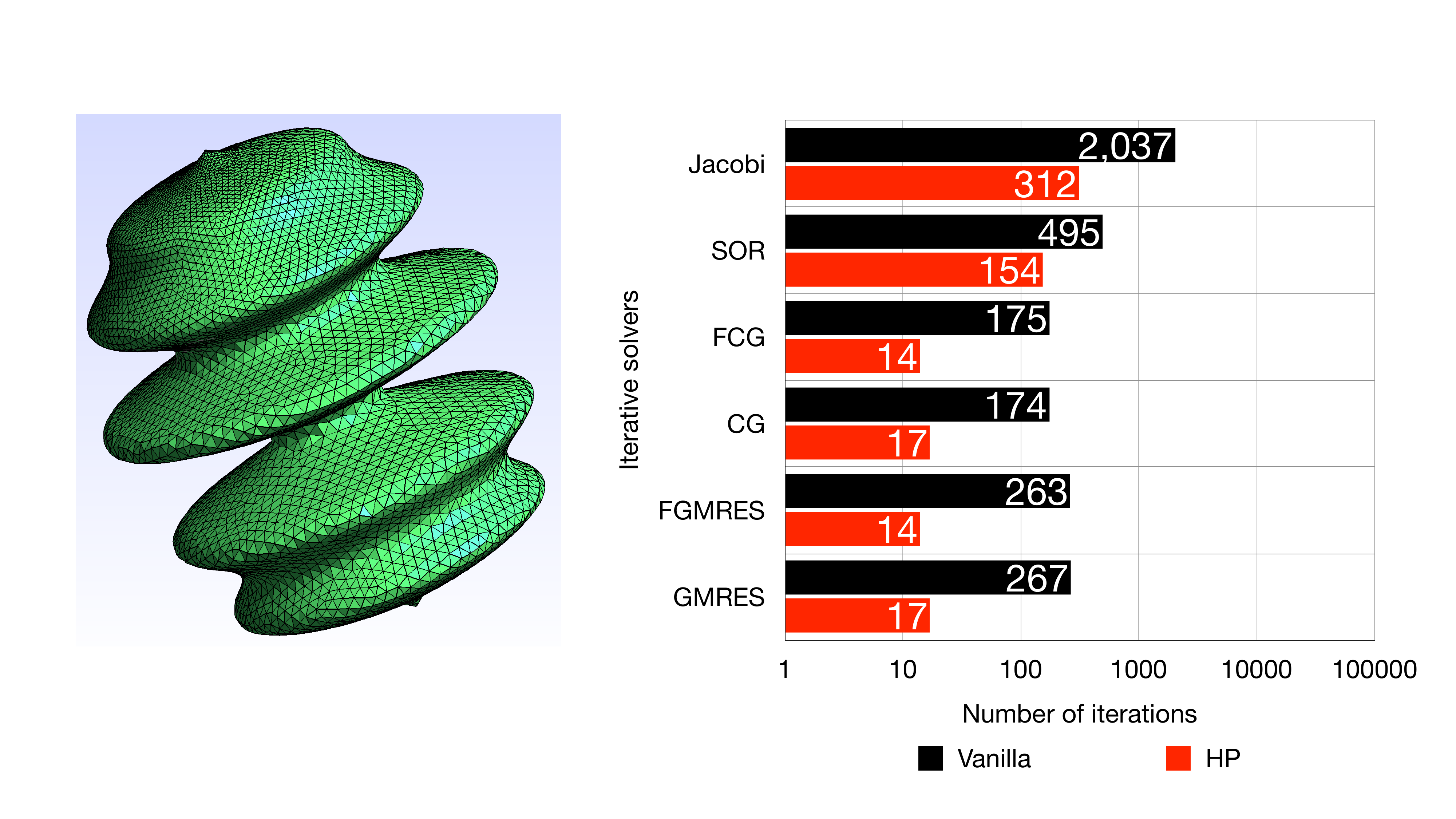}
	}
	\caption{The shape of the computational domains and the number of iterations required by the Relaxation methods~(Jacobi and SOR) and Krylov methods~(FCG, CG, FGMRES, and GMRES) to reach convergence. In the 2D problem, the maximum mesh size $h$, the number of relaxation step $n_{r}$, the number of trunk bases $k$ are set to $h=1/128$, $n_{r}=100$~(Relaxation) or $3$~(Krylov), and $k=30$, respectively.
	In the 3D problem, we set $h=1/32$, $n_{r}=10$~(Relaxation) or $7$~(Krylov), and $k=30$, respectively.
	Note that the TB approach in~\eqref{eqn:linearized_preconditioner} is applied for CG and GMRES, while the direct prediction of Geo-DeepONet is used otherwise.
	}\label{fig:comparison_poisson}
\end{figure}

\begin{figure}
	\centering
	\subfloat[Elasticity 2D]{
		\includegraphics[width=0.7\textwidth, trim={1cm, 2cm, 1cm, 4cm}, clip]{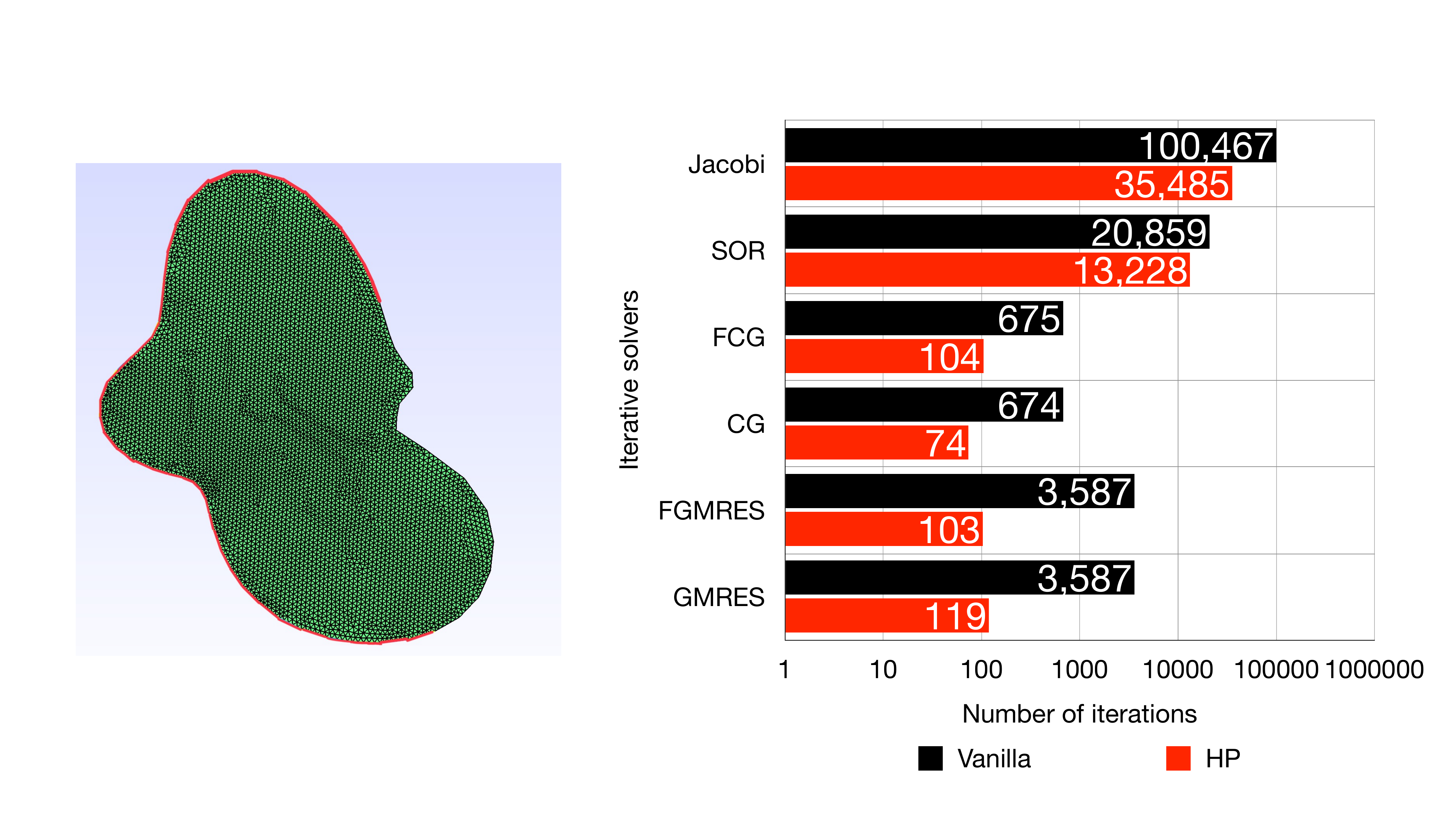}
	}
	\newline
	\subfloat[Elasticity 3D]{
		\includegraphics[width=0.7\textwidth, trim={1cm, 2cm, 1cm, 4cm}, clip]{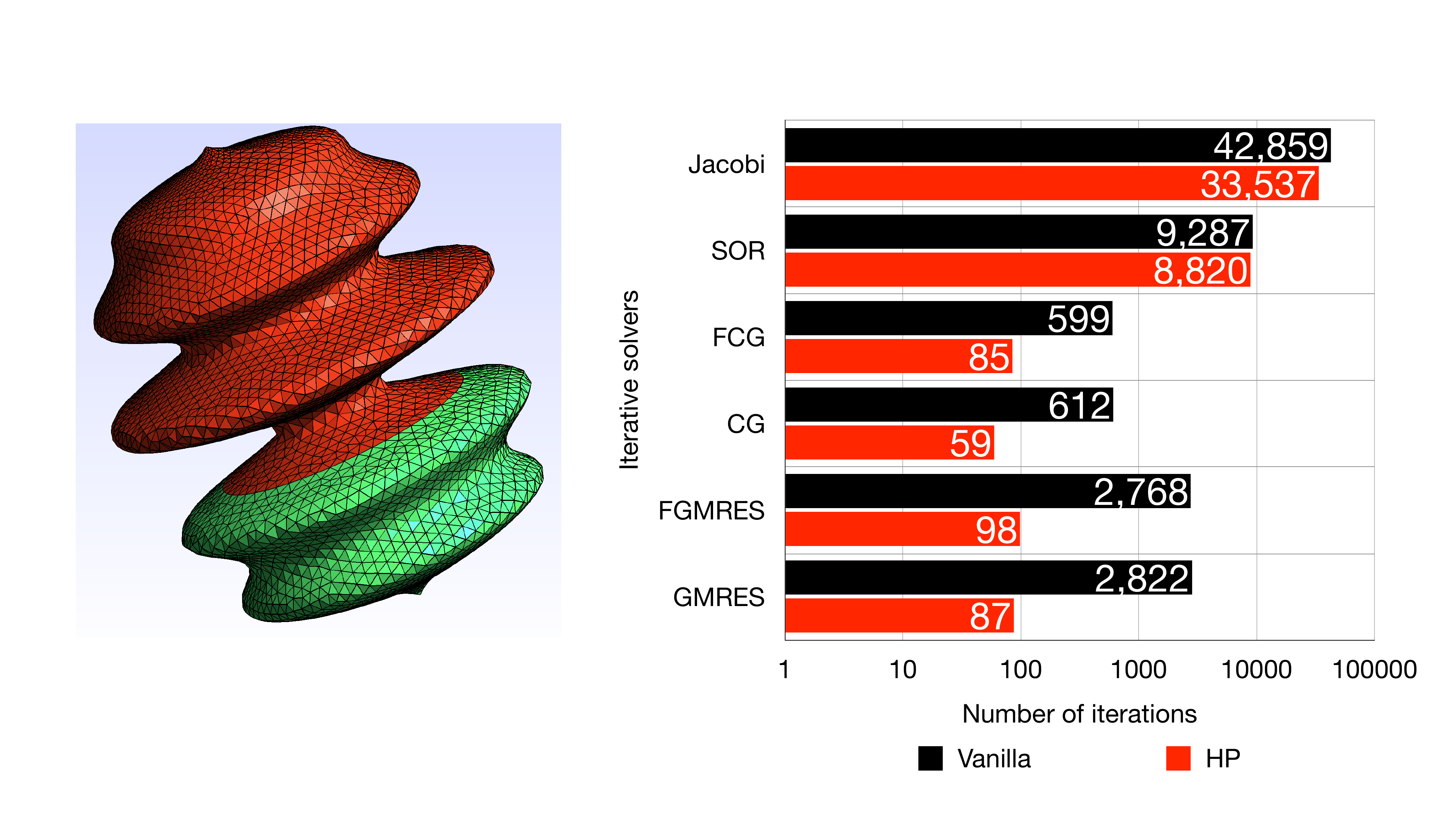}
	}
	\caption{The shape of the computational domains and the number of iterations required by the Relaxation methods~(Jacobi and SOR) and Krylov methods~(FCG, CG, FGMRES, and GMRES) to reach convergence.
		In each computational domain, Neumann boundary conditions are imposed on the red line in 2D (or the red region in 3D), while Dirichlet boundary conditions are imposed on the rest of the boundary.
		In the 2D problem, the maximum mesh size $h$, the number of relaxation step $n_{r}$, the number of trunk bases $k$ are set to $h=1/128$, $n_{r}=100$~(Relaxation) or $3$~(Krylov), and $k=30$, respectively.
		In the 3D problem, we set $h=1/32$, $n_{r}=50$~(Relaxation) or $3$~(Krylov), and $k=8$, respectively.
		Note that the TB approach in~\eqref{eqn:linearized_preconditioner} is applied for CG and GMRES, while the direct prediction of Geo-DeepONet is used otherwise.
	}\label{fig:comparison_elasticty}
\end{figure}

Next, we present the performance of the proposed hybrid preconditioned iterative solvers~(HP).
We use Jacobi, SOR, CG, FCG, GMRES, and FGMRES as backbone iterative solvers.
Note that CG and GMRES use the linear preconditioner constructed by TB approach while FCG and FGMRES use the nonlinear preconditioner consisting of the prediction of Geo-DeepONet.
We solve the Poisson equation and linear elasticity equation in 2D/3D unstructured domains.
In the chart of~\Cref{fig:comparison_poisson}, every HP iterative solver~(red) requires significantly less number of iterations than vanilla iterative solvers~(black).
The HP-relaxation methods~(Jacobi and SOR) achieve a speedup of approximately 7$\times$ on the benchmark problem.
As these methods are commonly used as smoothers for multigrid method, employing HP-relaxation method in this role can lead to faster convergence of the multigrid method~(see~\cite{kopanivcakova2024deeponet,lee2025fast,zhang2024blending}).
Furthermore, HP achieves a 10$\times$ or 20$\times$ acceleration in the CG or GMRES families, respectively, demonstrating its effectiveness across different iterative solvers.
The results of the linear elasticity equation, presented in~\Cref{fig:comparison_elasticty}, highlight a significant acceleration in the Krylov methods, even if Neumann boundary conditions are imposed on arbitrary boundary regions.
In particular, the GMRES families, which are commonly employed for solving nonsymmetric problems, experience an acceleration of approximately 30$\times$, demonstrating the capability of the HP approach in handling complex scenarios.

\subsection{Comparison with other preconditioners in real-world geometries}
\begin{figure}
	\centering
	\subfloat[Heat sink]{
		\includegraphics[width=0.3\textwidth]{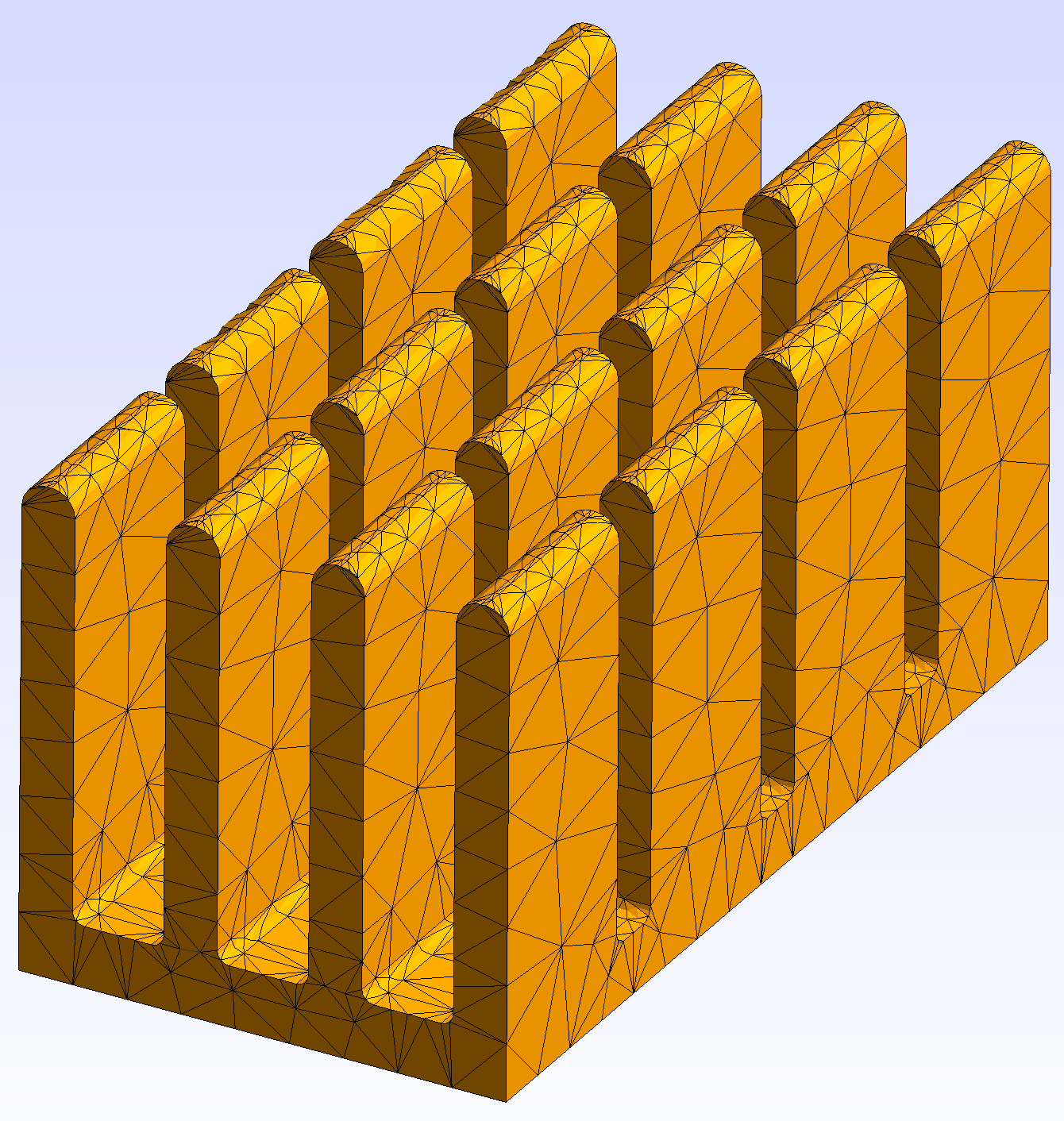}
	}
	\subfloat[Nut]{
		\includegraphics[width=0.3\textwidth]{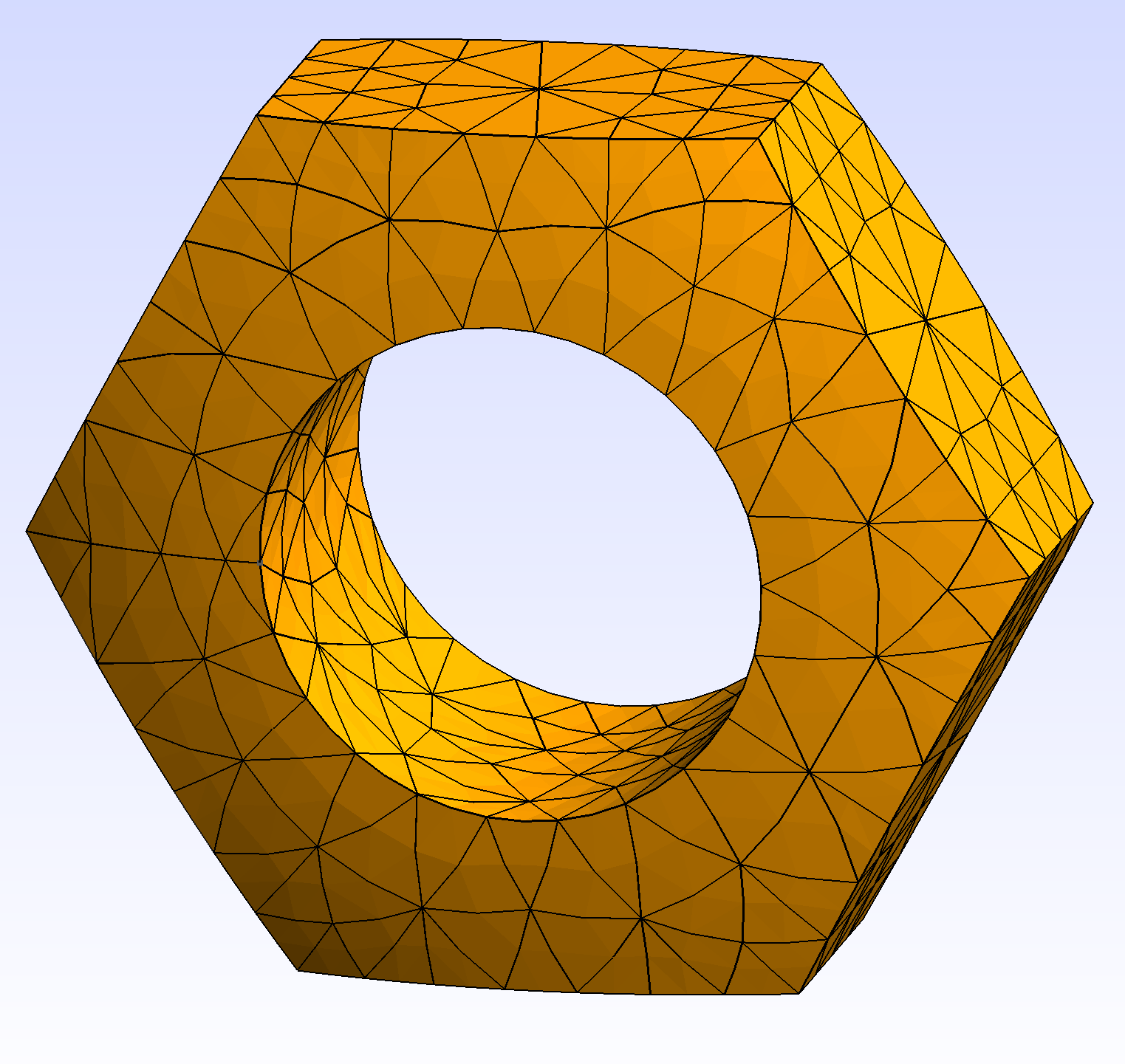}
	}
	\subfloat[Plastic embellisher]{
		\includegraphics[width=0.3\textwidth]{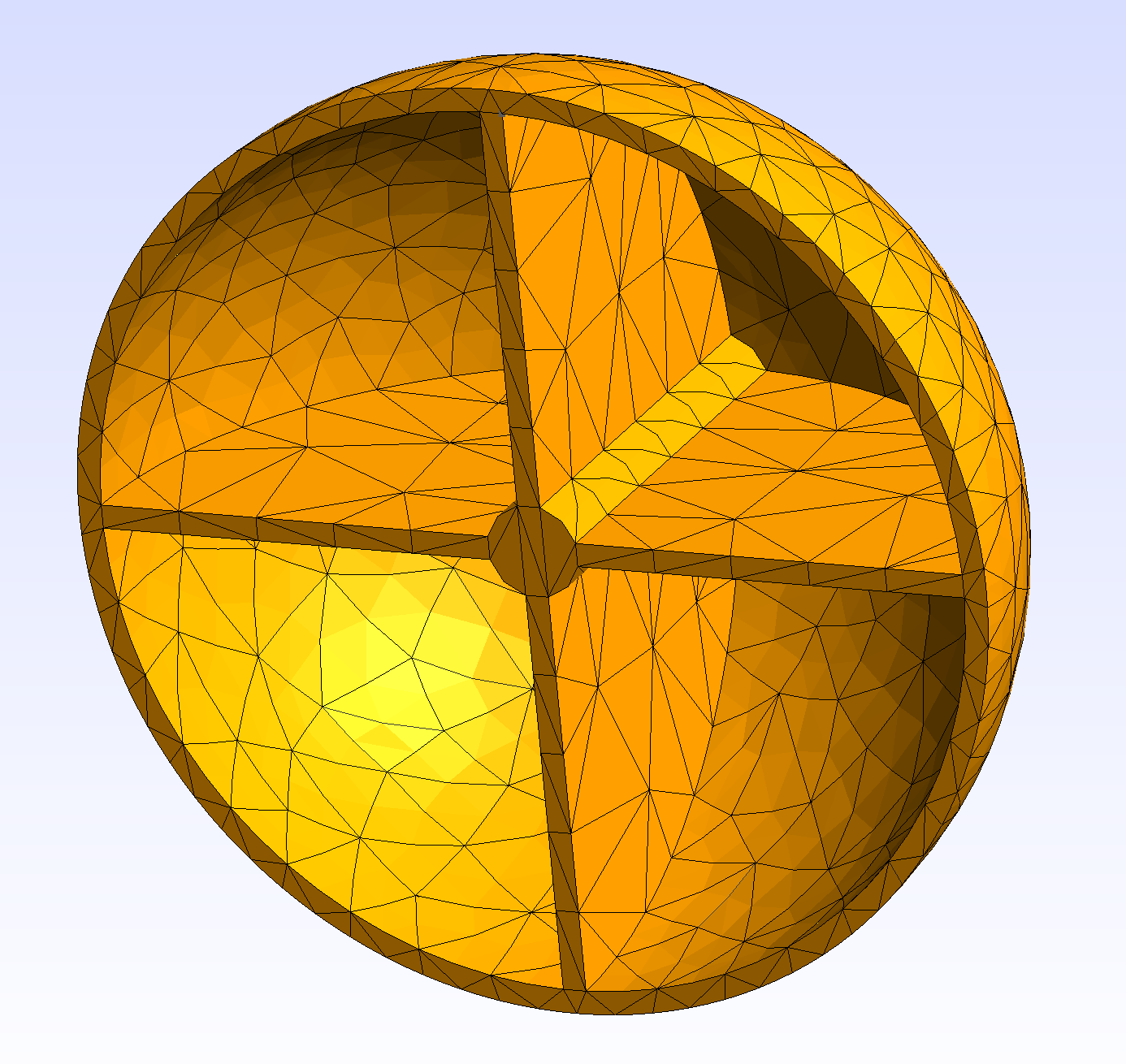}
	}
	\caption{Examples of meshes in 3D linear elasticity simulations. The $P^{1}$ conforming tetrahedron finite element is used for discretization.
		(a) The heat sink mesh consists of $13,824$ nodes and $7,178$ elements. Dirichlet boundary condition are imposed on the flat base, while Neumann boundary conditions are imposed elsewhere.
		(b) The nut mesh consists of $2,673$ nodes and $1,868$ elements. Dirichlet boundary conditions are imposed on the inner cylindrical surface, while Neumann boundary conditions are imposed elsewhere.
		(c) The plastic embellisher mesh consists of $3,717$ nodes and $2,289$ elements. Dirichlet boundary conditions are imposed on the flat base, while Neumann boundary conditions are imposed elsewhere.
	}\label{fig:real_geometry}
\end{figure}

\begin{figure}
	\centering
	\subfloat[Helmet]{
		\includegraphics[width=0.4\linewidth]{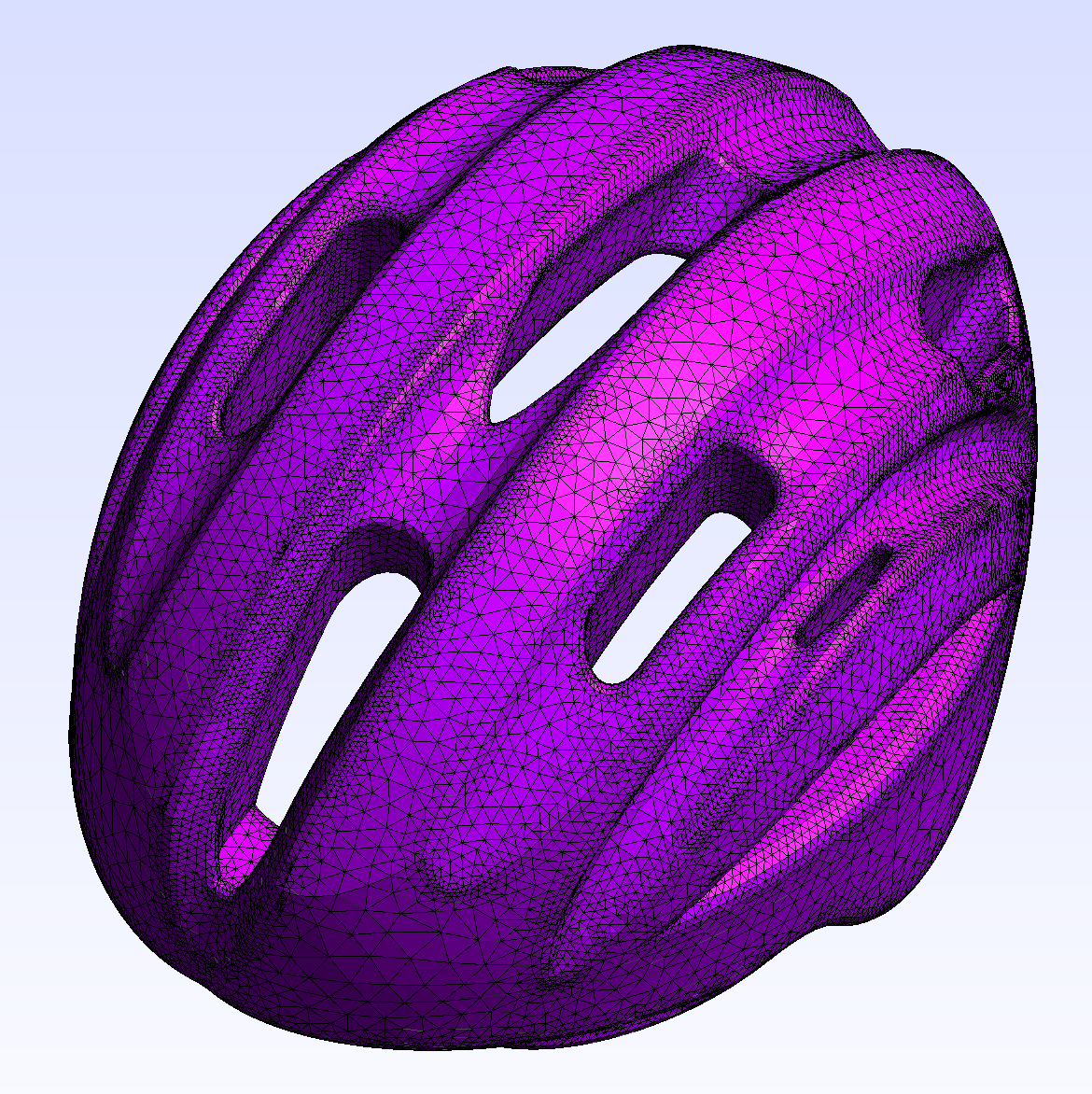}
	}
	\subfloat[Plate]{
		\includegraphics[width=0.4\linewidth]{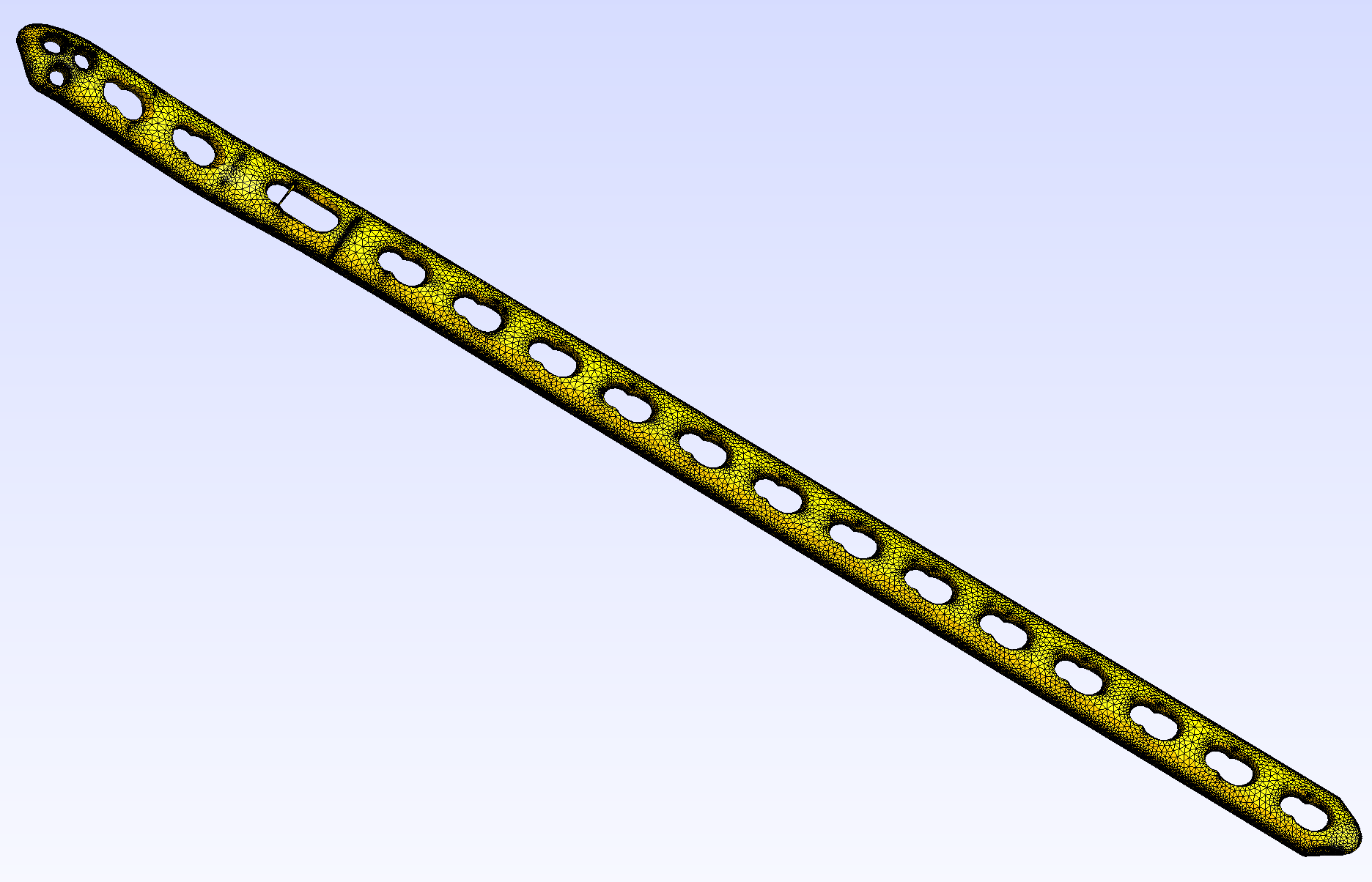}
	}
	\newline
	\subfloat[Frame]{
		\includegraphics[width=0.4\linewidth]{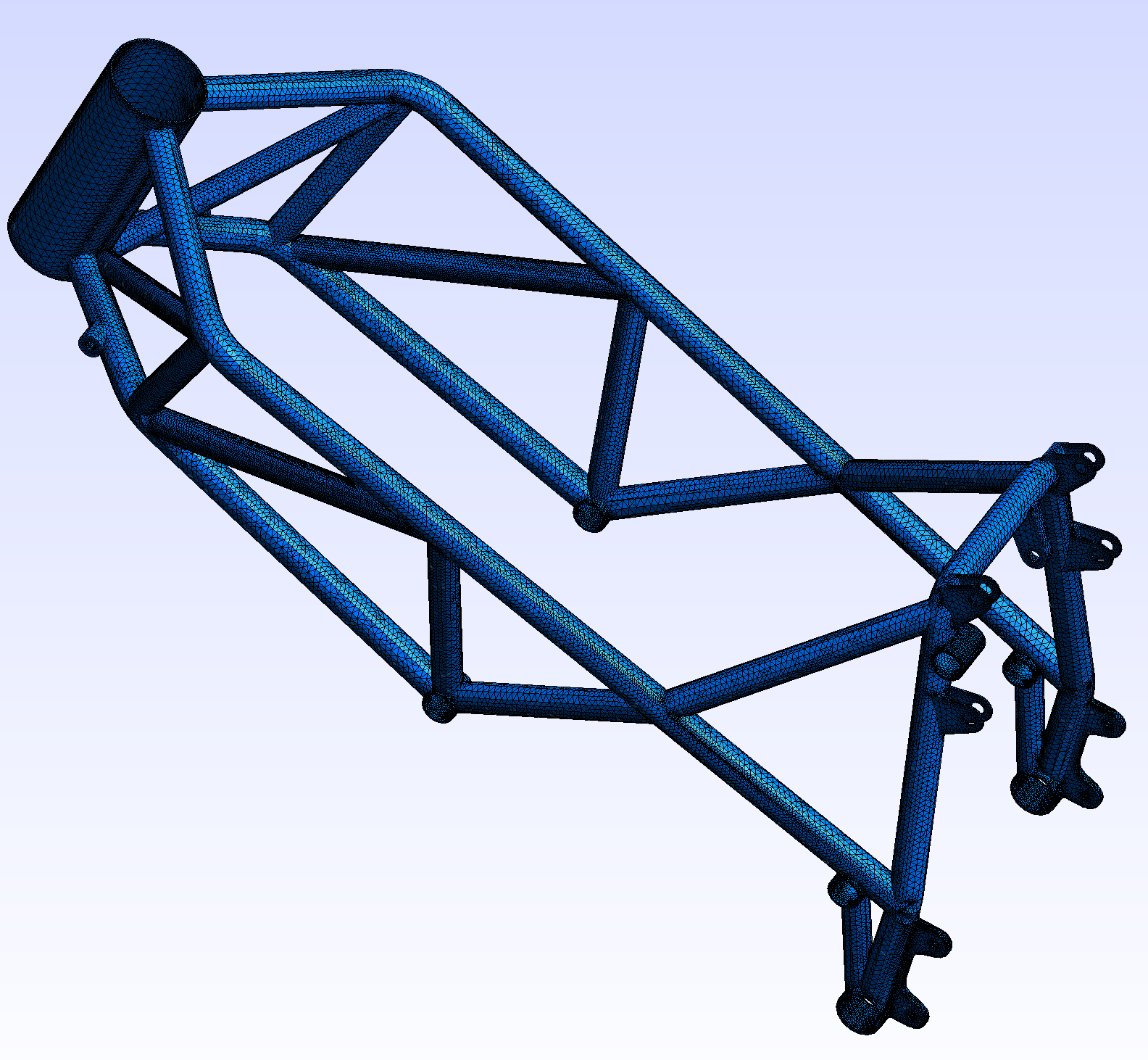}
	}
	\subfloat[Bracket]{
		\includegraphics[width=0.4\linewidth]{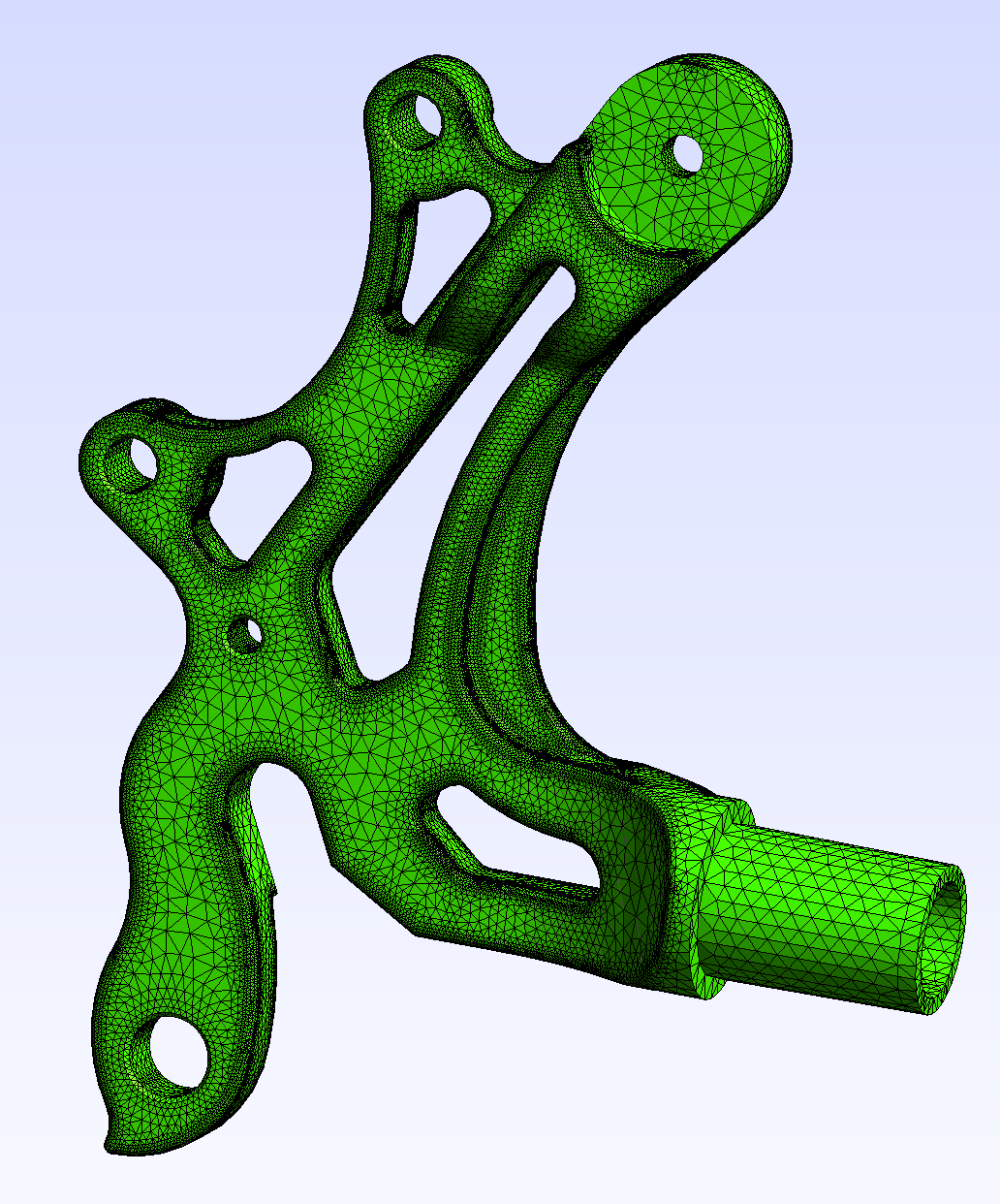}
	}
	\caption{Examples of meshes of 3D linear elasticity simulation. The $P^{1}$ conforming tetrahedron finite element is used for discretization.
		(a) The helmet mesh consists of $59,265$ nodes and $280,420$ elements. The Dirichlet boundary condition is imposed on the holes, while the Neumann boundary condition is imposed elsewhere.
		(b) The plate consists of $74,632$ nodes and $334,216$ elements. The Dirichlet boundary condition is imposed on the upper surface of plate excluding the center cylindrical hole, while the Neumann boundary condition is imposed elsewhere.
		(c) The frame mesh consists of $106,610$ nodes and $418,997$ elements. The Dirichlet boundary condition is imposed on the the main bars, while the Neumann boundary condition is imposed elsewhere.
		(d) The bracket mesh consists of $76,914$ nodes and $340,333$ elements. The Dirichlet boundary condition is imposed on the backside of the bracket and the connecting parts, while the Neumann boundary condition is imposed elsewhere.
	}\label{fig:real_geometry_large}
\end{figure}

Finally, we compare the performance of the proposed geometry-aware hybrid preconditioner~(HP) with other commonly used preconditioners.
We solve the 3D linear elasticity equation in various real-world meshes, which are presented in~\Cref{fig:real_geometry}.
To setup the HP iterative solvers, we reuse the parameters of the Geo-DeepONet trained in surfaces of revolution and utilize the Chebyshev semi-iterative method~\cite{adams2003parallel} for the relaxation steps.
Note that each mesh has a different number of nodes, elements, and positioning of Dirichlet boundary conditions.
We employ the CG method as the backbone iterative solver and utilize the ILU and SOR preconditioners implemented in the PETSc library and the preconditioner implemented in the software ANSYS Mechanical.
\Cref{tab:comparison_real} presents the results obtained for the heat sink, nut, and plastic embellisher meshes.
The HP-CG shows the best performance and achieves about 16$\times$ acceleration compared to the vanilla CG, while the ILU preconditioner in PETSc sometimes fails to converge.
In particular, HP demonstrates significantly larger speedup compared to other preconditioners when the computational domain has a symmetric shape, such as the nut and the plastic embellisher.
This advantage can be attributed to the fact that Geo-DeepONet was trained on randomly generated surfaces of revolution, which essentially capture the characteristics of such symmetric geometries.
Training Geo-DeepONet on a more diverse dataset consisting of various CAD models would enhance its robustness and generalizability, leading to even more consistent performance improvements across a wider range of geometries, such as the heat sink.
Additionally, we investigate the performance of the proposed HP iterative solvers when the number of unknowns is greater than $50,000$, which is presented in~\Cref{fig:real_geometry_large}.
Algebraic multigrid~(AMGs) is a commonly used preconditioner to achieve algorithmic scalability of CG.\@
Here, we use smoothed aggregation algebraic multigrid~(SA-AMG)~\cite{vanek1996algebraic} in PETSc and BoomerAMG~\cite{yang2002boomeramg} in HYPRE~\cite{falgout2002hypre}.
We also combine the proposed hybrid preconditioning framework with AMG. Specifically, we can define the projected trunk bases using the restriction and prolongation operators defined in each AMG level and combine it with classical smoothers.
\Cref{tab:comparison_amg} shows that the hybrid preconditioned solvers outperform the baseline solvers and successfully reduce the number of iterations required to solve the given problem.
The numerical results of additional real-world geometries are presented in~\Cref{sec:additional}.

\begin{table}
	\centering
	\caption{The number of iterations required by the preconditioned CG method solving 3D linear elasticity equation in various real world meshes.
		For HP, we set the number of relaxation steps and the number of trunk bases to $n_{r}=3$ and $k=8$, respectively. Note that the ILU preconditioner failed to converged for the heat sink and plastic embellisher cases.
        }\label{tab:comparison_real}
	\begin{tabular}{lccc}
		\toprule
		Preconditioner type & Heat sink   & Nut        & Plastic embellisher \\
		\midrule\midrule
		None                & 1,152       & 77         & 365                 \\
		\midrule
		ILU~(PETSc)         & ---         & 18         & ---                 \\
		\midrule
		SOR~(PETSc)         & 300         & 22         & 89                  \\
		\midrule
		HP~(Ours)           & \textbf{71} & \textbf{5} & \textbf{20}         \\
		\midrule
		ANSYS               & 123         & 34         & 117                 \\
		\bottomrule
	\end{tabular}
\end{table}

\begin{table}
	\centering
	\caption{The number of iterations required by the preconditioned CG method solving 3D linear elasticity equation in various large scale real world meshes.
		For HP, we set the number of relaxation steps and the number of trunk bases to $n_{r}=3$ and $k=8$, respectively. For HP-SA-AMG, we set $n_{r}=1$.}\label{tab:comparison_amg}
	\begin{tabular}{lcccc}
		\toprule
		Preconditioner type & Helmet       & Plate       & Frame        & Bracket      \\
		\midrule\midrule
		SOR~(PETSc)         & 10,350       & 38          & 8,594        & 2,953        \\
		\midrule
		HP~(Ours)           & 4,255        & 15          & 3,513        & 1,203        \\
		\midrule\midrule
		BoomerAMG~(HYPRE)   & 402 & 19          & 372          & \textbf{100} \\
		\midrule
		SA-AMG~(PETSc)      & 1,030        & 22          & 807          & 471          \\
		\midrule
		HP-SA-AMG~(Ours)    & \textbf{367}          & \textbf{12} & \textbf{229} & 140          \\
		\bottomrule
	\end{tabular}
\end{table}

\section{Conclusion}\label{sec:conclusion}
In this paper we proposed geometry-aware hybrid preconditioned iterative solvers for parametric partial differential equations in arbitrary unstructured domains.
Hybridization was performed using the multiplicative subspace correction framework, blending standard iterative methods with neural operators.
In order to address the arbitrary geometries, we proposed a Geo-DeepONet that utilizes a Chebyshev spectral graph convolutional network and a scaling network.
Leveraging the graph structure of the discretized domain, the coordinates and the adjacency matrix were used as inputs for the proposed Geo-DeepONet.
The performance of the proposed hybrid preconditioned iterative solver was evaluated using benchmark problems.

Despite the promising results, a few limitations remain in the current framework.
First, the Geo-DeepONet was trained on datasets generated from closed curves or surfaces of revolution.
To ensure robust generalization to more complex scenarios, future studies should extend the training dataset to include arbitrary real world geometries, such as CAD geometries.
Second, the proposed Geo-DeepONet involves projecting feature values from unstructured mesh nodes onto a structured mesh.
This process can become computationally intensive, particularly when the domain is discretized with a highly refined mesh.
Future work will address this latter shortcoming by designing a fully node-based network architecture such as the Point transformer~\cite{zhao2021point}.
Furthermore, if the size of the domain is very large, domain decomposition methods are commonly used.
FETI-DP~\cite{farhat2001feti} or BDDC~\cite{mandel2003convergence} preconditioners, known as SOTA preconditioners, require solving an interface problem defined on the interfaces of subdomains.
However, the condition number of the problem could be large when the original problem is ill-conditioned.
To address this challenge, the eigenvalue problem is traditionally solved on the interface to construct the augmented coarse space~\cite{bootland2023overlapping}, which has the difficulty of having to be solved anew when the domain changes. This issue could be handled by utilizing the proposed Geo-DeepONet to construct the augmented coarse space.

\section*{Acknowledgments}
Y.L. was partially supported in part by Basic Science Research Program through NRF funded by the Ministry of Education (No.~RS2023--00247199).
G.E.K. is supported by the ONR Vannevar Bush Faculty Fellowship. We also acknowledge support from the DOE-MMICS SEA-CROGS DE-SC0023191 award and Ansys Inc.

\appendix
\bmsection{Hybrid Iterative Numerical Transferable Solver}\label{sec:hints}
In this section we provide practical details of the hybrid iterative numerical transferable solver~(HINTS) framework~\cite{zhang2024blending}.
Let us recall that the iteration of HINTS can be written as
\begin{equation*}
	\begin{split}
		\rv^{(i)}     & = \fv - \Km\uv^{(i)},                         \\
		\uv^{(i+1/2)} & = \uv^{(i)} + \M_{1}^{-1}(\rv^{(i)}),         \\
		\rv^{(i+1/2)} & = \fv - \Km\uv^{(i+1/2)},                     \\
		\uv^{(i+1)}   & = \uv^{(i+1/2)} + \M_{2}^{-1}(\rv^{(i+1/2)}), \\
	\end{split}
\end{equation*}
where the symbols $\M_{1}^{-1}$ and $\M_{2}^{-1}$ denote the $n_{r}$ iterations of the relaxation method and the prediction of the neural operator, respectively.
For example, if we choose the Jacobi method as the relaxation method and $n_{r}=10$, the HINTS-Jacobi will run $10$ iterations of the Jacobi method and then apply the neural operator.

The length of the intermediate residual vector $\rv^{(i+1/2)}$ depends on the discretization of the domain, which can be different from the size of input required for the neural operator prediction.
For example, assume the pre-trained neural operator takes as input a function discretized in the 2D unit square domain with mesh size of $h=1/30$, while the given 2D unit square domain is discretized with a maximum mesh size of $h=1/128$.
Then, the size of the input function for the neural operator should be $31^{2}=961$, while the size of the intermediate residual vector is $129^{2} = 16,641$.
To input this residual vector into the neural operator, a projection operator from the finite element space to the neural operator space is required, which can be done using linear, quadratic, or cubic interpolation.
Let us denote this projection as $\Pi_{h}$.
Finally, given the residual vector $\rv$ and the coordinate vector $\xv \in \R^{n \times d}$, the second preconditioner $\M_{2}^{-1}$ is defined as
\begin{equation*}
	\M_{2}^{-1}(\rv) = \G_{\text{NN}}(\Pi_{h}(\rv))(\xv),
\end{equation*}
where $\G_{\text{NN}}$ is the pre-trained neural operator.
Note that the symbols $n$ and $d$ denote the number of nodes and the geometric dimension of the domain.

\bmsection{Trunk basis hybridization approach}\label{sec:tb}
In this section, we provide practical details of the TB approach proposed in~\cite{kopanivcakova2024deeponet}.
For a given user-specified number $k$ and pre-trained DeepONet, the $(i,j)$-th component of the representaion matrix of TB based prolongation operator $\Pm \in \R^{n \times k}$ is given by
\begin{equation*}
	{[\Pm]}_{ij} = T_{j}(\xv_{i}),
\end{equation*}
where $T_{j}(\xv_{j})$ is the $j$-th component of the output of the trunk network evaluated at the node point $\xv_{j} \in \Omega \subset \R^{d}$.
Note that the symbols $n$ and $d$ denote the number of nodes and the geometric dimension of the domain.
To define a well-conditioned coarse-level operator $\Km_{c}:=\Pm^{\ast} \Km \Pm$, it is essential to ensure that the operator $\Pm$ has full rank and orthogonal columns.
This can be easily addressed by performing the QR decomposition of $\Pm$ and using $\Qm$ as the prolongation operator instead of $\Pm$.

In addition, the selection of trunk basis functions is another crucial part that affects the condition number of $\Km_{c}$.
Several selection methods were suggested in~\cite{kopanivcakova2024deeponet}, including selecting in order, randomly, or based on smaller singular values.
In this work, we select the trunk basis function in order.

\bmsection{Details of training the neural operator}\label{sec:homogeneous}
In this section we claim that the neural operator doesn't need to be trained on problems with nonhomogeneous Dirichlet boundary conditions when it is used as the backbone of the hybrid preconditioning framework.

Recall that the second preconditioner defined in~\eqref{eqn:hybrid} acts on the intermediate residual $\rv^{(i+1/2)}$ at each iteration, not the right-hand side vector $\fv$.
Without loss of generality, let us assume that we are solving the following boundary value problem:
\begin{equation}
	\label{eqn:model}
	\left\{
	\begin{aligned}
		- u''(x)                         & =f \text{ in } \Omega = (0,1), \\
		u(0)                             & = g_{1}(0),                    \\
		\frac{\partial u}{\partial n}(1) & = g_{2}(1).
	\end{aligned}
	\right.
\end{equation}
Under the finite element discretization, the interval $\Omega = (0,1)$ is equidistantly discretized as
\begin{equation*}
	0=x_{0}< x_{1}< x_{2} < x_{3} < x_{4}=1.
\end{equation*}
Here, the length of each subinterval is $h=1/4$.
Then, the discretized linear system of the problem~\eqref{eqn:model} is formulated as
\begin{align}
	\label{eqn:matrixsystem}
	\underbrace{
		\begin{bmatrix}
			1  & 0  & 0  & 0  & 0  \\
			-1 & 2  & -1 & 0  & 0  \\
			0  & -1 & 2  & -1 & 0  \\
			0  & 0  & -1 & 2  & -1 \\
			0  & 0  & 0  & -1 & 1  \\
		\end{bmatrix}
	}_{\Am}
	\underbrace{
		\begin{bmatrix}
			u_{0} \\
			u_{1} \\
			u_{2} \\
			u_{3} \\
			u_{4} \\
		\end{bmatrix}
	}_{\uv}
	 & =
	\underbrace{
		\begin{bmatrix}
			g_{1}(x_0)              \\
			h^{2}f(x_1)             \\
			h^{2}f(x_2)             \\
			h^{2}f(x_3)             \\
			h^{2}f(x_4)-hg_{2}(x_4) \\
		\end{bmatrix}
	}_{\fv}.
\end{align}
From the formulation~\eqref{eqn:matrixsystem}, the boundary values~$\{u_{0}, u_{4}\}$ are determined by
\begin{equation*}
	\left\{
	\begin{aligned}
		u_0 & = g_{1}(x_0),                      \\
		u_4 & = h^{2}f(x_4) - hg_{2}(x_4) + u_3.
	\end{aligned}
	\right.
\end{equation*}
When we utilize any iterative solver, the $i$-th residual vector $\rv^{(i)}$ is given by
\begin{equation*}
	\rv^{(i)} = \fv - \Am\uv^{(i)} =
	\begin{bmatrix}
		g_{1}(x_0) - u_0^{(i)} = 0                                   \\
		h^{2}f(x_{1}) - (-u_{0}^{(i)} + 2 u_{1}^{(i)} - u_{2}^{(i)}) \\
		h^{2}f(x_{2}) - (-u_{1}^{(i)} + 2 u_{2}^{(i)} - u_{3}^{(i)}) \\
		h^{2}f(x_{3}) - (-u_{2}^{(i)} + 2 u_{3}^{(i)} - u_{4}^{(i)}) \\
		h^{2}f(x_4)-h g_{2}(x_4) + u_{3}^{(i)} - u_{4}^{(i)} = 0     \\
	\end{bmatrix}.
\end{equation*}
Since the computed residual vector $\rv^{(i)}$ is zero on the boundary, independently of $g_{1}$ and $g_{2}$, it is sufficient for neural operators to be trained on problems with nonhomogeneous Dirichlet boundary conditions.
This property eliminates the need to generate the boundary functions such as $g_{1}$ and $g_{2}$ for configuring training samples, thereby reducing the memory requirements and simplifying the training process.

\begin{figure}
	\centering
	\subfloat[Poisson 2D~(Circle with hole)]{
		\includegraphics[width=0.6\textwidth, trim={1cm, 0cm, 0cm, 4cm}, clip]{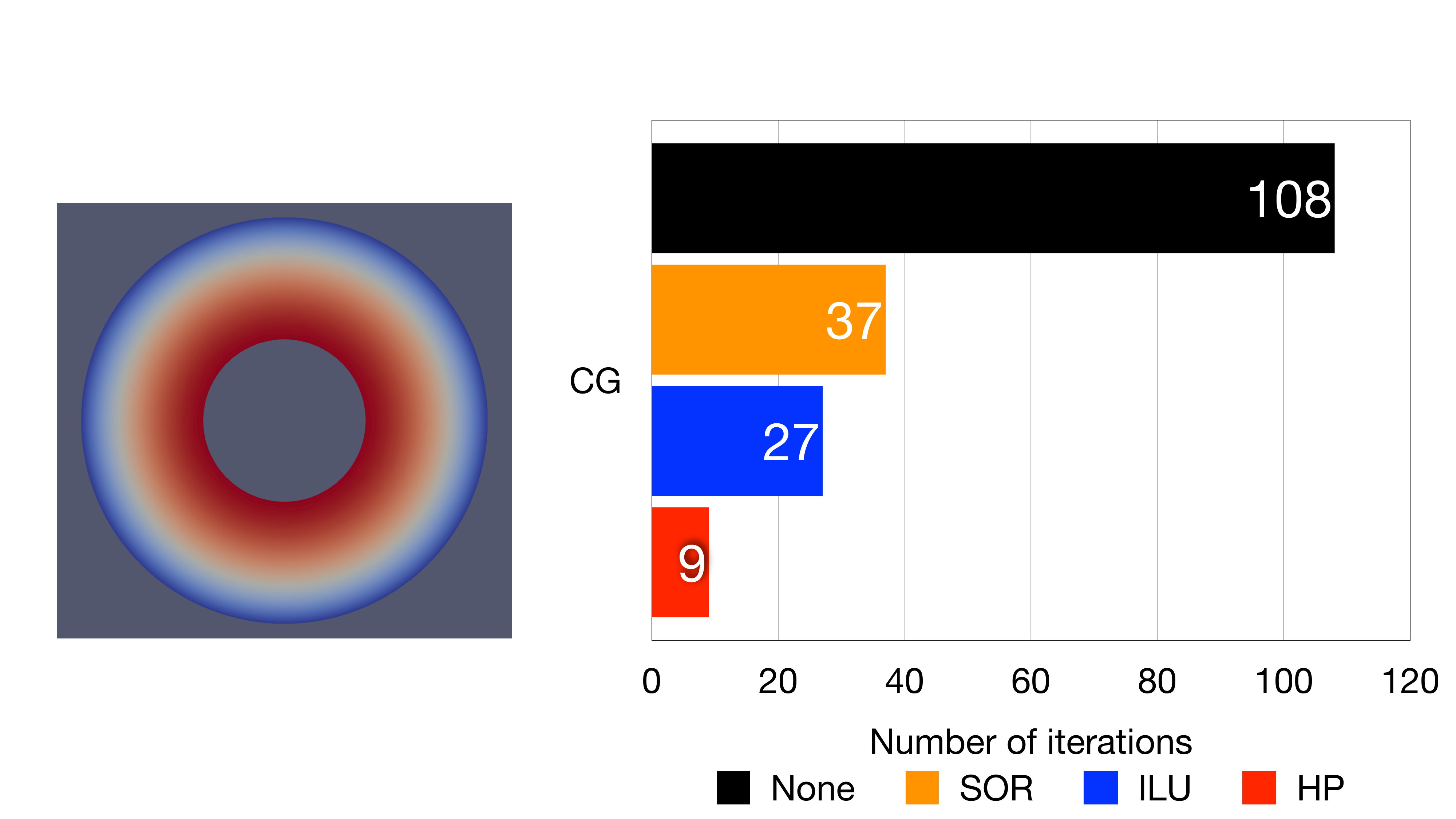}
	}
	\newline
	\subfloat[Elasticity 2D~(Wrench)]{
		\includegraphics[width=0.6\textwidth, trim={1cm, 0cm, 0cm, 4cm}, clip]{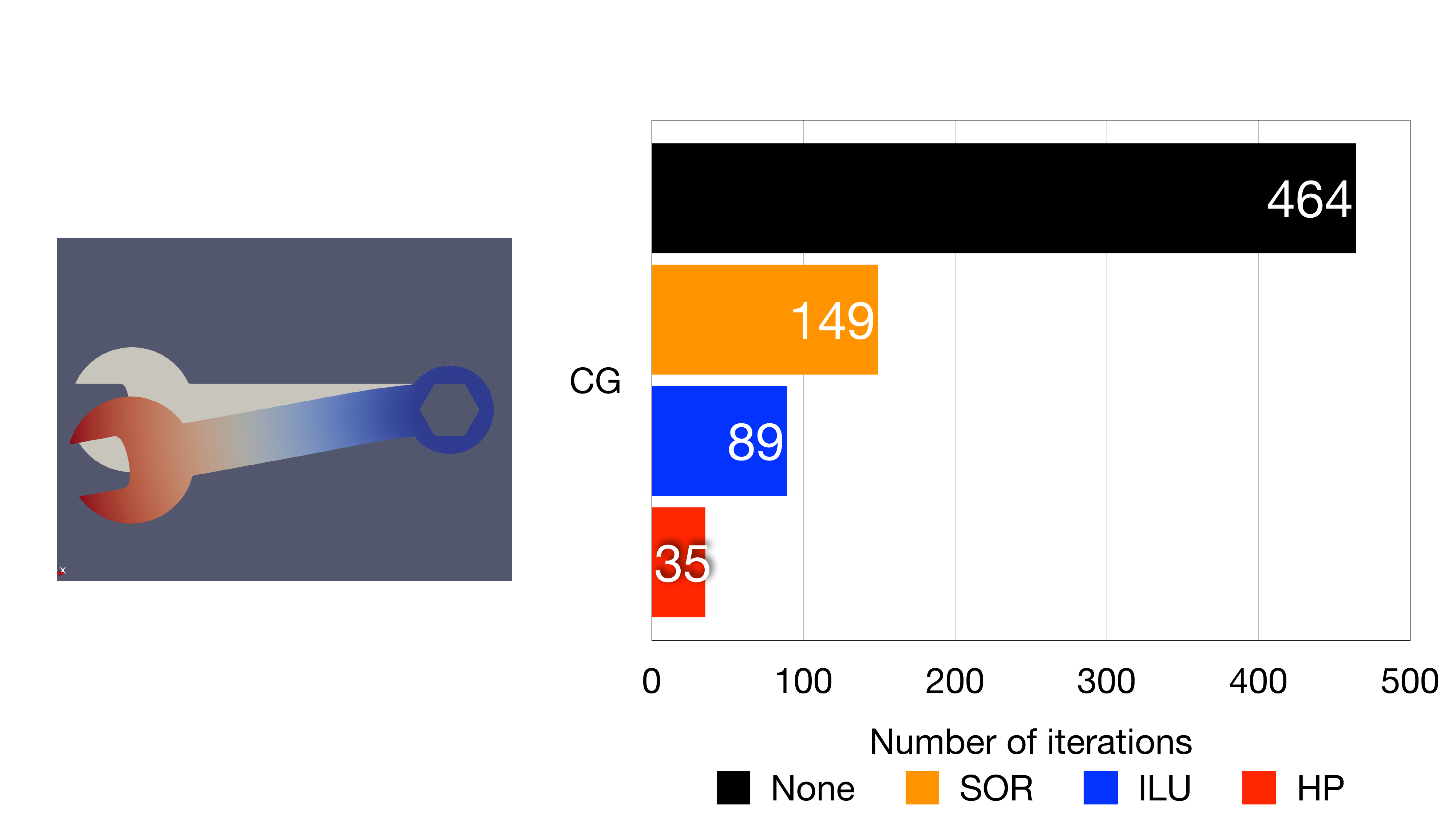}
	}
	\newline
	\subfloat[Elasticity 3D~(E shape)]{
		\includegraphics[width=0.6\textwidth, trim={1cm, 0cm, 0cm, 4cm}, clip]{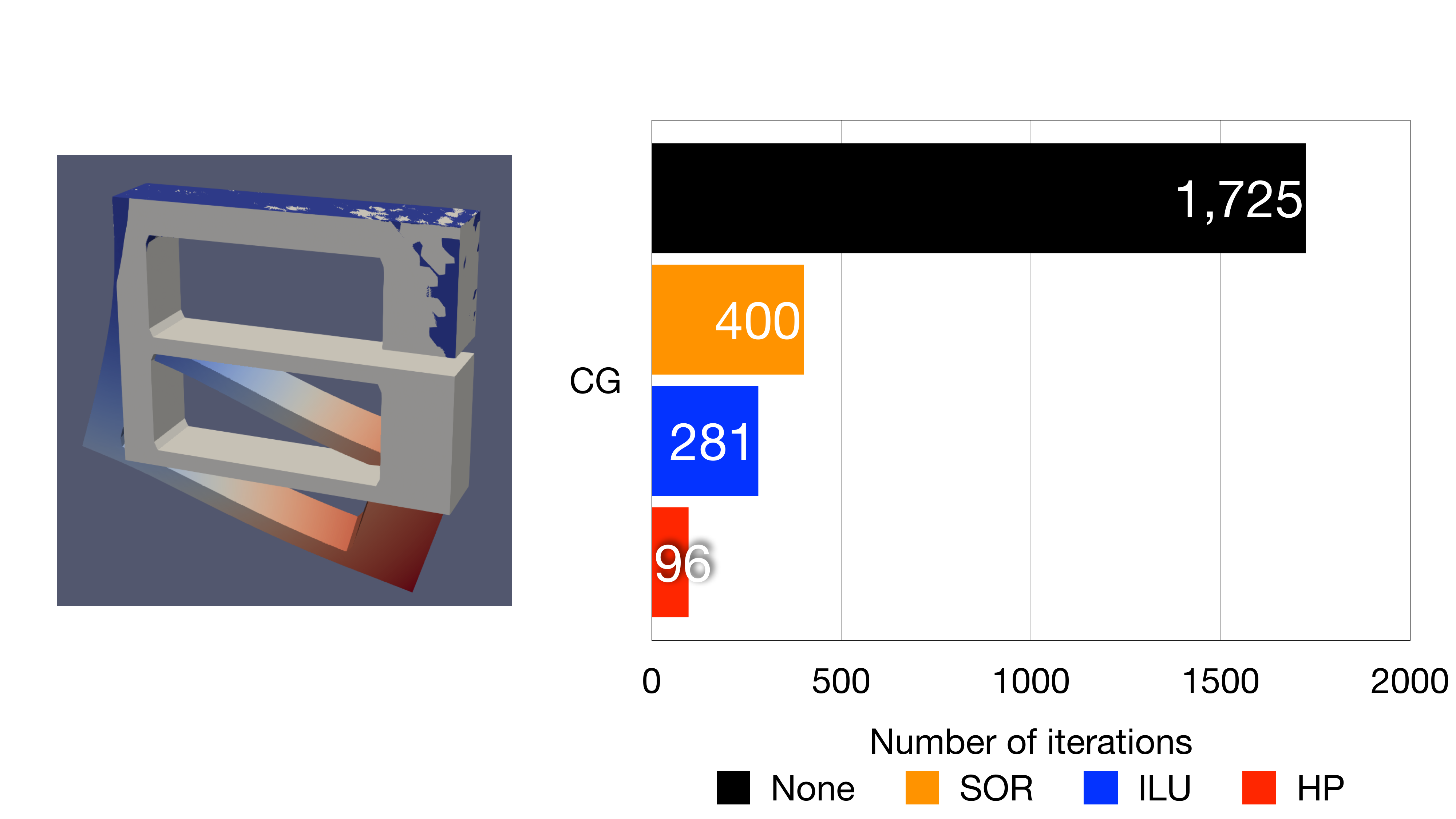}
	}
	\caption{The shape of the computational domains with corresponding solutions and the number of iterations required by the conjugate gradient method~(CG) to reach convergence.
		In each geometry, Dirichlet boundary conditions are imposed on the outer circle, inner hexagon, and top part, respectively.
		Note that Neumann boundary conditions are imposed on the rest of the boundary.
		In the elasticity problem, the gray and color shape indicates the given geometry and the corresponding solution, respectively.
		The hybridization ratio $n_{r}$ and the number of trunk bases $k$ are set to $n_{r}=7$ and $k=8$, respectively.
	}\label{fig:additonal}
\end{figure}

\bmsection{Additional numerical results}\label{sec:additional}
In this section we present additional numerical results of the proposed geometry-aware hybrid preconditioner~(HP).
We use the conjugate gradient method~(CG), the most popular iterative method for solving elliptic equations, with various preconditioners commonly used in real-world geometries.
In~\Cref{fig:additonal}, we can observe that the HP shows much better performance than SOR and ILU without additional training.
Note that the Geo-DeepONet used for HP is pre-trained with arbitrary generated curves~(2D) or surfaces of revolution~(3D).

\bibliography{reference}

\end{document}